\documentclass[11pt]{article}
\usepackage{acl}
\usepackage{latexsym}
\usepackage[T1]{fontenc}
\usepackage[utf8]{inputenc}
\usepackage{microtype}
\usepackage{placeins}
\usepackage{graphicx}
\usepackage{booktabs}
\usepackage{amsmath}
\usepackage{enumitem}
\usepackage{adjustbox}
\usepackage{url}
\usepackage{iftex} 
\ifPDFTeX
  \usepackage[T1]{fontenc}
  \usepackage{times}
  \newcommand{\hinditxt}[1]{#1}
  \newcommand{\bengalitxt}[1]{#1}
  \newcommand{\tamiltxt}[1]{#1}
  \newcommand{\telugutxt}[1]{#1}
\else
  \usepackage{fontspec}
  \usepackage[english]{babel}

  \babelprovide[import, main]{english}
  \babelprovide[import, onchar=ids fonts]{hindi}
  \babelprovide[import, onchar=ids fonts]{bengali}
  \babelprovide[import, onchar=ids fonts]{tamil}
  \babelprovide[import, onchar=ids fonts]{telugu}

  \babelfont{rm}[BoldFont=texgyretermes-bold.otf,ItalicFont=texgyretermes-italic.otf,BoldItalicFont=texgyretermes-bolditalic.otf]{texgyretermes-regular.otf}

  \babelfont[*devanagari]{rm}[Path=fonts/,Scale=MatchLowercase,BoldFont=NotoSerifDevanagari-Regular.ttf,ItalicFont=NotoSerifDevanagari-Regular.ttf]{NotoSerifDevanagari-Regular.ttf}
  \babelfont[*bengali]{rm}[Path=fonts/,Scale=MatchLowercase,BoldFont=NotoSerifBengali-Regular.ttf,ItalicFont=NotoSerifBengali-Regular.ttf]{NotoSerifBengali-Regular.ttf}
  \babelfont[*tamil]{rm}[Path=fonts/,Scale=MatchLowercase,BoldFont=NotoSerifTamil-Regular.ttf,ItalicFont=NotoSerifTamil-Regular.ttf]{NotoSerifTamil-Regular.ttf}
  \babelfont[*telugu]{rm}[Path=fonts/,Scale=MatchLowercase,BoldFont=NotoSerifTelugu-Regular.ttf,ItalicFont=NotoSerifTelugu-Regular.ttf]{NotoSerifTelugu-Regular.ttf}

  \newcommand{\hinditxt}[1]{\foreignlanguage{hindi}{#1}}
  \newcommand{\bengalitxt}[1]{\foreignlanguage{bengali}{#1}}
  \newcommand{\tamiltxt}[1]{\foreignlanguage{tamil}{#1}}
  \newcommand{\telugutxt}[1]{\foreignlanguage{telugu}{#1}}
\fi
\title{When Ground Truth Is Not Ground Truth: Automatic Labeling for Hallucination Detection}
\author{Rounak Sen \And Divya Kumari \And Deeksha Varshney}
\begin{document}
\maketitle

\begin{abstract}
Hallucinations in Large Language Models (LLMs) remain a major challenge, as they can produce misleading, factually incorrect, or unverifiable information, thereby undermining their trustworthiness and reliability. 
However, reliable ground-truth labels remain a key challenge in evaluating hallucination detectors. Recent studies often construct labels automatically by comparing model generations with reference answers using similarity metrics such as BLEURT. This implicitly treats BLEURT similarity as a proxy for hallucination, despite its English-centric supervision, unbounded score range, and design as a similarity metric rather than a factuality measure, making threshold-based ground-truth construction potentially unreliable.
Through comprehensive experiments using 15 automatic supervision judges, seven hallucination detection baselines, two generator models, two QA datasets, and five languages, we demonstrate that detector performance varies substantially with the choice of supervision judge. Human validation reveals substantial differences in labeling reliability, with QGen-9B achieving the lowest overall misclassification rate (12.86\%), while BLEURT and BLEURT-n exhibit rates of 47.23\% and 47.32\%, respectively. Our findings highlight the limitations of automatic similarity-based labeling and the need for more reliable judging methods that better capture semantic and factual correctness.




\end{abstract}

\section{Introduction}
Hallucinations are a persistent limitation of large language models (LLMs), where fluent generations may contain misleading, unsupported, or factually incorrect content \cite{ji2023survey, kaddour2023challenges, xu2024hallucination,zhang2025siren,alansari2026large}. As LLMs are increasingly deployed in applications where reliability is important, detecting such generations has become an active area of research \cite{varshney2023stitch, dhuliawala2023chain, chuang2024dola, shi-etal-2024-trusting}. With the rapid development and adoption of multilingual and culturally diverse LLMs, hallucination detection has also expanded beyond English, raising additional challenges in reliably identifying hallucinations across languages \cite{alansari2026crosshallu,alvi2026multihaludet}.

The use of automatic labeling strategies has become increasingly common in hallucination benchmarks and detector evaluations \cite{janiak2025illusion, du2024haloscope, zhang2026probe, turpin2023language, bhatnagar2026drift, zhang2025icr, burns2023discovering}. A common strategy is to compare a generated response with a reference using an automatic similarity metric and convert the resulting score into a binary label. Although attractive for its scalability, this procedure introduces an important assumption that is rarely examined: \textit{does similarity to a reference provide a reliable indication of whether a generation is hallucinated ?}

\begin{figure}[t]
    \centering
    \includegraphics[width=0.49\textwidth]{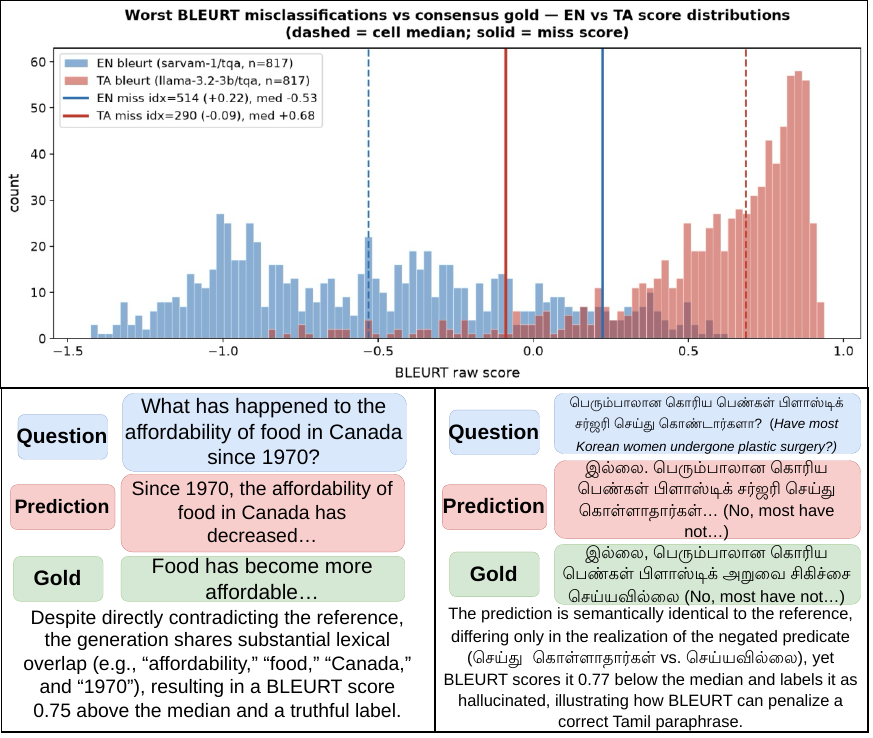}
    \caption{Two complementary misclassification cases: high similarity labels a contradictory English response as truthful, while low similarity labels a semantically correct Tamil paraphrase as hallucinated.}
    \label{fig:motivation}
\end{figure}

Recent independent evidence strengthens this concern.~\citet{janiak2025illusion} show that replacing lexical evaluation with an LLM-based judge can substantially change the reported performance of established hallucination detectors, with drops of up to 45.9\% for some methods. ~\citet{chen2026rethinking} explicitly identify label noise as a major challenge in hallucination-detection benchmarks and show that noisy labels can hinder detector performance. HaloScope  \citep{du2024haloscope}, provides an implementation that uses BLEURT or ROUGE to determine whether generated answers are truthful, with a threshold controlling the resulting labels. Similarly, TSV \citep{park2025tsv} evaluates a generation as truthful when its similarity to the reference exceeds a predefined BLEURT threshold. Once this procedure is adopted, the similarity metric is no longer merely an auxiliary evaluation measure: its errors become ground-truth errors. A detector evaluated against such labels can therefore appear effective while learning to separate BLEURT-defined categories rather than hallucinated from non-hallucinated generations. 

As illustrated in Figure~\ref{fig:motivation}, BLEURT exhibits substantially different score distributions across languages, making a universal threshold difficult to justify. More importantly, the figure highlights two contrasting failure cases. The English response by Sarvam-1 (index 514) receives a BLEURT score of $+0.22$, substantially above the language-specific median of $-0.53$, and is therefore labeled as \textit{truthful} by BLEURT. However, the generated response directly contradicts the reference: it states that food affordability in Canada has decreased since 1970, whereas the reference states that food has become more affordable. The high score is largely driven by lexical overlap between the generation and reference, including ''affordability,'' ''food,'' ''Canada,'' and ''1970.'' In contrast, the Tamil response by Llama-3.2-3B (index 290) receives a BLEURT score of $-0.09$, substantially below the language-specific median of $+0.68$, and is consequently labeled as \textit{hallucinated}. However, human evaluation unanimously identifies the response as correct, with the generation and reference expressing the same meaning and differing only in linguistic realization. These examples expose two opposing failure modes: BLEURT can assign high similarity to a contradictory generation while assigning low similarity to a semantically correct multilingual paraphrase. Thus, thresholding BLEURT can transform differences in lexical overlap and language-specific realization into incorrect ground-truth labels.

To establish a fair comparison, we evaluate multiple generator models, supervision judges, datasets, languages, and detector families under consistent data splits.

Specifically, we make two contributions:
\begin{enumerate}[nolistsep]
    \item We systematically evaluate 15 automatic supervision judges and seven hallucination detectors across two QA datasets and five languages, demonstrating how the choice of supervision labels influences measured detection performance.

    \item We validate automatic labels against human annotations and identify systematic failure modes across supervision methods, highlighting the limitations of similarity-based labeling and the need for independently validated, semantically grounded supervision.
\end{enumerate}


\section{Related Work}

Hallucinations remain a persistent limitation of large language models (LLMs), which can generate fluent and coherent responses that are factually inaccurate, unsupported by evidence, or inconsistent with the input context \citep{zhang2025siren,alansari2026large}. Hallucination detection methods can broadly be categorized into supervised and unsupervised approaches. Supervised methods train classifiers on labeled examples of truthful and hallucinated generations, often leveraging the internal representations of LLMs \citep{azaria2023internal,orgad2025llms,arteaga2024hallucination}. These methods range from probing model activations, as in SAPLMA~\citep{azaria2023internal}, to analyzing inter-layer transitions, as in DRIFT~\citep{bhatnagar2026drift}.

In contrast, unsupervised methods detect potential hallucinations without requiring labeled training examples, relying instead on signals such as token-level uncertainty derived from model probabilities \citep{ren2022out,kuhn2023semantic}, semantic disagreement across multiple generations \citep{malinin2020uncertainty,farquhar2024entropy}, and patterns in internal representations \citep{chen2024inside,sriramanan2024llm}. Representative approaches include sampling-based consistency checks, such as SelfCheckGPT~\citep{selfcheckgpt}, which requires multiple generations per query; representation-based methods, including subspace analysis in INSIDE~\citep{chen2024inside} and hidden-state attribution in HaloScope~\citep{du2024haloscope}; and methods that examine specific architectural components, such as ICR-Probe's comparison of multi-head self-attention and feed-forward network contributions~\citep{zhang2025icr} and prompt-embedding probes~\citep{mrykhin2026pep}. Steering-based approaches, notably Truthful Subspace Valuation (TSV)~\citep{park2025tsv}, further exploit internal activation directions to assess generation reliability and provide a foundation for our detector family. However, constructing ground-truth labels at scale remains challenging, motivating the use of automatic labeling strategies that compare generated responses with reference answers using similarity metrics.

Recent work has extended hallucination detection to multilingual settings. CrossHallu examine Arabic-English transfer and report that performance depends on class separability and language alignment~\citep{alansari2026crosshallu}, while MultiHaluDet probes hidden-state signals across French, Bengali, and Amharic~\citep{alvi2026multihaludet}. Islam et al.~\citep{islam2025much} investigate multilingual hallucination estimation across 30 languages by transferring a detector trained on English data, while Rykov et al.~\citep{rykov2025models} introduce PsiloQA, a multilingual dataset for span-level hallucination detection across 14 languages. Other resources include Poly-FEVER~\citep{zhang2025poly}, a multilingual fact-verification benchmark, and BHRAM-IL~\citep{terdalkar2025bhram}, which evaluates hallucination recognition across several Indian languages. However, these rely on translation or transfer from English, potentially limiting their ability to capture language-specific expressions and cultural contexts.

\textit{The Illusion of Progress}~\citep{janiak2025illusion}, which re-audits QA hallucination detection and reports limited agreement between ROUGE-based reference-overlap labels and human annotations. Although \cite{janiak2025illusion} showed limitations, recent hallucination-detection pipelines still employ learned similarity metrics such as BLEURT to automatically assign truthfulness labels~\citep{du2024haloscope,park2025tsv}. Hence, we analyze BLEURT-based labeling across languages and generator models.

\begin{table*}[h!]
\centering
\small
\setlength{\tabcolsep}{2.2pt}
\renewcommand{\arraystretch}{0.92}
\caption{\textbf{Comprehensive detector benchmark across all fifteen supervision judges.} Test AUROC averaged across $2$ generator models, $5$ languages, and $2$ QA datasets from the full benchmark grid. Best score across judges for each detector (row-wise) is bold.}
\label{tab:methods}
\begin{adjustbox}{max width=\textwidth}
\begin{tabular}{@{}l|ccc|cc|ccc|cccc|ccc@{}}
\toprule
& \multicolumn{3}{c|}{\textbf{Learned Similarity}} & \multicolumn{2}{c|}{\textbf{Surface Overlap}} & \multicolumn{3}{c|}{\textbf{Textual Entailment}} & \multicolumn{4}{c|}{\textbf{Generative Verdicts}} & \multicolumn{3}{c}{\textbf{Decision Agents}} \\
\textbf{Baseline Method} & BLEURT & \shortstack{BLEURT\\-n} & \shortstack{BLEURT\\-20} & COMET & \shortstack{ROUGE\\-L} & NLI & \shortstack{Mini\\Check} & \shortstack{FAct\\Score} & \shortstack{QGen\\-4B} & \shortstack{QGen\\-9B} & \shortstack{Gemma\\Gen} & Rubric & JevK5 & SemIf & Laya \\
\midrule
DRIFT & \textbf{0.725} & \textbf{0.725} & \textbf{0.823} & \textbf{0.688} & \textbf{0.737} & \textbf{0.580} & \textbf{0.795} & \textbf{0.581} & \textbf{0.640} & \textbf{0.636} & 0.612 & \textbf{0.577} & \textbf{0.615} & \textbf{0.640} & \textbf{0.666} \\
SAPLMA & 0.708 & 0.708 & 0.787 & 0.657 & 0.696 & 0.551 & 0.793 & 0.563 & 0.613 & 0.612 & 0.584 & 0.574 & 0.591 & 0.611 & 0.623 \\
ICR-magnitude & 0.602 & 0.601 & 0.337 & 0.557 & 0.506 & 0.483 & 0.325 & 0.533 & 0.453 & 0.513 & 0.453 & 0.491 & 0.477 & 0.498 & 0.429 \\
HaloScope-subspace & 0.653 & 0.653 & 0.345 & 0.589 & 0.551 & 0.473 & 0.252 & 0.530 & 0.468 & 0.528 & 0.442 & 0.479 & 0.476 & 0.511 & 0.402 \\
\shortstack[l]{Length-Normalized\\Predictive Entropy} & 0.404 & 0.403 & 0.678 & 0.446 & 0.425 & 0.551 & 0.727 & 0.534 & 0.619 & 0.549 & \textbf{0.623} & 0.531 & 0.601 & 0.599 & 0.659 \\
\shortstack[l]{Answer Log-\\Probability} & 0.524 & 0.523 & 0.496 & 0.509 & 0.446 & 0.528 & 0.495 & 0.532 & 0.525 & 0.531 & 0.556 & 0.536 & 0.518 & 0.524 & 0.559 \\
\shortstack[l]{Answer Length\\Control} & 0.693 & 0.693 & 0.345 & 0.591 & 0.701 & 0.474 & 0.270 & 0.505 & 0.464 & 0.531 & 0.426 & 0.503 & 0.438 & 0.519 & 0.369 \\
\bottomrule
\end{tabular}
\end{adjustbox}
\end{table*}

\section{Experimental Setup}
\label{sec:setup}

To investigate how automatic supervision influences hallucination detection, we construct alternative training label sets using
fifteen automatic supervision judges. For each judge, we apply its labeling procedure to the training set, producing a corresponding
set of labels for the generated responses. We then use these judge-specific labels to train and evaluate seven baseline hallucination detection methods. The experiments consider two
base generator models, five languages, and two QA datasets.
\paragraph{Datasets and languages.} We evaluate on TruthfulQA~\citep{lin2022truthfulqa} and SciQ~\citep{welbl2017crowdsourcing}, in English, Hindi, Bengali, Tamil, and Telugu. The train and test sets contain 612, 205 and 750, 250 QA pairs, respectively. More details and translation-quality metrics are in Appendix~\ref{app:transqa}. 

\paragraph{Generator Models} We evaluate two open-weight LLMs: Llama-3.2-3B~\citep{llama3modelcard} and Sarvam-1~\citep{sarvam1modelcard}, 3B model.

\paragraph{Supervision judges.}  We examine fifteen automated labeling methods spanning five technical mechanisms. Learned similarity methods include BLEURT~\citep{bleurt}, BLEURT-n, and BLEURT-20~\citep{pu2021learning}. Surface-overlap methods include COMET~\citep{rei2020comet}, ROUGE-L~\citep{lin2004rouge}, with these methods excluded from benchmark rankings due to saturation. Textual entailment methods comprise NLI~\citep{he2023debertav3}, MiniCheck~\citep{tang2024minicheck}, and FActScore~\citep{min2023factscore}. Generative verdict methods include QGen-4B~\citep{qwen35modelcard}, QGen-9B~\citep{qwen35modelcard}, GemmaGen~\citep{gemma3report}, and Rubric~\citep{gemma3report}. Finally, decision-agent methods comprise Laya~\citep{layamodel}, JevK5~\citep{jevk5model}, and SemIf~\citep{lee2026semif}. {All methods evaluate candidate responses against the gold reference set using max-over-references aggregation, defined as $\max_{r \in R} \text{Score}(c, r)$ (except NLI, which averages the entailment margin over $R$, and FActScore, defined at atomic fact level). More details in Appendix~\ref{app:judges16}.
}

\paragraph{Baseline Hallucination Detection methods.} We evaluate seven hallucination detectors spanning four methodological families. Supervised hidden-state probes include DRIFT~\citep{bhatnagar2026drift}, which tracks inter-layer representation transitions, and SAPLMA~\citep{azaria2023internal}, which uses final-token activations. Unsupervised activation-based methods include HaloScope-subspace\footnote{\scriptsize Implements only the class-conditional subspace-distance scoring at fixed depth taps; the membership-estimation half is not reimplemented hence \textit{HaloScope-subspace}, never \textit{HaloScope}.}~\citep{du2024haloscope} and ICR-magnitude\footnote{\scriptsize Implements only the representation-dynamics signal of the mean per-layer rate of change $\|h_{l+1}-h_l\| / \|h_l\|$ without the paper's learned MHSA-vs-FFN attribution head; hence \textit{ICR-magnitude}, not \textit{ICR}.}~\citep{zhang2025icr}. Output-based uncertainty baselines comprise Length-Normalized Predictive Entropy~\citep{farquhar2024entropy} and Answer Log-Probability~\citep{kuhn2023semantic}. Finally, Answer Length Control provides a non-parametric baseline based on response length~\citep{janiak2025illusion}.

\paragraph{Evaluation Methods}
We use AUROC \cite{Bradley1997TheUO} as the primary measure of hallucination performance. 




\paragraph{Implementation Details}
We evaluated against single-sample representatives of hallucination detection methods, as multi-sample approaches incur substantial inference overhead and are not budget-comparable to a single forward pass. We adopt median splits for unbounded metrics (BLEURT) and standard 0.5 cutoffs for bounded metrics to mirror standard benchmark practices \citep{du2024haloscope,park2025tsv}. We note that median splitting imposes a balanced prior, conflating ranking discrimination with prevalence calibration; see Appendix~\ref{app:ablations} for threshold stability analyses. All model generation, judge scoring and linear detector probe training were executed on a server equipped with 2$\times$ NVIDIA RTX 6000 Ada GPUs (48GB VRAM each, 96GB aggregate), an AMD EPYC 9124 16-Core Processor, and 256GB RAM. An anonymized repository containing our complete codebase, prompt templates, and evaluation pipeline is available on github\footnote{\url{https://anonymous.4open.science/r/judge_hal-3305/}}. Detailed implementation and hyperparameter configurations are provided in Appendix~\ref{app:ablations}.

\begin{figure*}[h!]
    \centering
    \includegraphics[width=\linewidth]{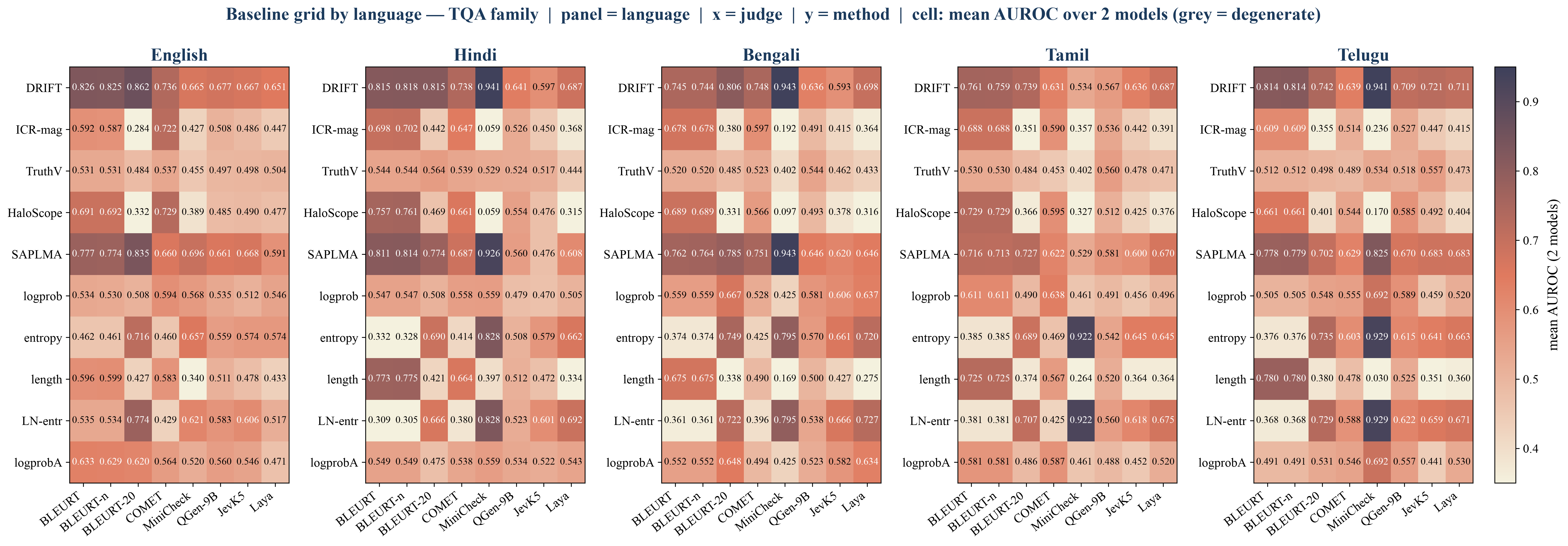}
    \caption{{Language-wise AUROC on TruthfulQA for eight supervision judges, averaged across two generator models. Panels correspond to English, Hindi, Bengali, Tamil, and Telugu. The results illustrate language-dependent variation in detector performance across supervision judges.}}
    \label{fig:gridlang_tqa}
\end{figure*}

\section{Results and Analysis}

\subsection{Preliminary Analysis}

We first examine how the choice of supervision judge affects the measured performance of the seven baseline hallucination detectors. We report detector performance averaged across the five languages, two QA datasets, and two generator models to obtain an aggregate view of the benchmark. Table~\ref{tab:methods} presents the resulting AUROC scores for seven baseline hallucination detection methods under fifteen supervision judges.

Across the learned-similarity judges, DRIFT achieves AUROCs of $0.725$ under both BLEURT and BLEURT-n, and $0.823$ under
BLEURT-20. SAPLMA follows a similar pattern, with AUROCs of $0.708$, $0.708$, and $0.787$, respectively. Under surface-overlap
judges, DRIFT achieves $0.688$ with COMET and $0.737$ with ROUGE-L, while the Answer Length Control baseline reaches $0.591$ and $0.701$, respectively. Performance varies under textual-entailment judges. DRIFT achieves an AUROC of $0.795$ with MiniCheck, compared with $0.580$ with NLI and $0.581$ with FActScore. SAPLMA exhibits a similar advantage under MiniCheck ($0.793$), whereas ICR-magnitude and HaloScope-subspace perform substantially worse under this judge, with AUROCs of $0.325$ and $0.252$, respectively. 

Under generative-verdict judges, DRIFT achieves AUROCs of $0.640$ and $0.636$ with QGen-4B and QGen-9B, respectively,
and $0.612$ with GemmaGen. Length-Normalized Predictive Entropy achieves $0.619$ under QGen-4B and $0.623$ under GemmaGen, exceeding DRIFT in the latter case. Under Rubric, DRIFT reaches
$0.577$, while SAPLMA achieves $0.574$. Finally, among the decision-agent judges, DRIFT obtains AUROCs of $0.615$ with
JevK5, $0.640$ with SemIf, and $0.666$ with Laya. SAPLMA achieves corresponding scores of $0.591$, $0.611$, and $0.623$.

Figure~\ref{fig:gridlang_tqa} presents baseline AUROC across five
languages on TruthfulQA, averaged over the two generator models, for eight supervision judges. The results reveal substantial
language-wise variation, indicating that no judge consistently achieves the highest discrimination across all languages. Although
BLEURT-20 is designed for multilingual evaluation, its performance is not uniformly superior. For DRIFT, BLEURT-20 achieves the
highest AUROC among the displayed judges in English ($0.862$) and Hindi ($0.815$), whereas MiniCheck yields substantially higher
scores in Bengali ($0.943$) and Telugu ($0.941$). In Tamil, BLEURT and BLEURT-n achieve comparable AUROCs of $0.761$ and
$0.759$, respectively, while MiniCheck reaches only $0.534$. This contrast suggests that multilingual design alone does not
ensure consistent discrimination across languages. Similar variation is observed for the generative judges. Although the QGen judges are based on Qwen models, their performance varies across languages and detector families rather than consistently
outperforming the other judges. For example, under DRIFT, QGen-4B achieves AUROCs of $0.641$ in Hindi, $0.636$ in Bengali, and
$0.567$ in Tamil, while MiniCheck reaches $0.941$, $0.943$, and $0.534$ in these languages, respectively. The contrast is
particularly pronounced in Hindi and Bengali, where MiniCheck provides substantially stronger discrimination than QGen-4B. Results on SciQ are provided in
Appendix~\ref{app:detailed_baselines}.

These aggregate and language-wise variations raise a central question: do similarity-based judges provide reliable supervision for hallucination detection, or do they favor detectors that reproduce their own biases ? Section~\ref{sec:f2_transfer} computes cross-ground-truth transferability across 15 judges.
We also complement the quantitative results with qualitative examples to examine cases where similarity-based labels conflict with semantic correctness, further investigating this question in Section~\ref{sec:qualitative_val}.



\subsection{Cross-Ground-Truth Transferability}
\label{sec:f2_transfer}

Our preceding experiments show that measured detector performance varies with the choice of supervision judge. To investigate whether different judges provide consistent supervision signals, we extend the baseline evaluation with a cross-ground-truth transfer experiment, using DRIFT and SAPLMA as a representative supervised detector. We train them separately using the binary labels produced by each of the fifteen judges and evaluate each trained probe against the labels of every judge on the same held-out test examples. For each ordered pair of training and evaluation judges $(j_a,j_b)$, the probe is trained on the training split labeled by $j_a$, its weights are frozen, and its predictions are evaluated against the test labels from $j_b$ using AUROC. Representations, data splits, and detector architecture remain fixed across comparisons, allowing us to examine how changing the supervision source affects measured performance. The resulting $15\times15$ matrix is shown in Figure~\ref{fig:cross_gt_matrix}. Results for SAPLMA is shown in Appendix \ref{app:xgt_full}.

Figure~\ref{fig:cross_gt_matrix} reveals a substantial difference between in-judge performance and cross-judge transfer. The diagonal entries achieve a mean AUROC of $0.669$, whereas the off-diagonal entries average $0.530$, close to chance ($0.500$). Defining the transfer gap as $\Delta(j_a,j_b)=A(j_a,j_a)-A(j_a,j_b)$, where $A(j_a,j_b)$ denotes the AUROC of a probe trained on judge $j_a$ and evaluated against judge $j_b$, the difference between the mean diagonal and off-diagonal scores is $0.139$. For example, DRIFT trained
using BLEURT labels achieves an AUROC of $0.729$ when evaluated
against BLEURT labels, but only $0.360$ against BLEURT-20 labels.
Conversely, training and evaluating with BLEURT-20 yields an AUROC
of $0.822$. These differences indicate that detector performance
is sensitive to the supervision and evaluation criteria, rather
than being directly comparable across judges.

\begin{figure}[h!]
    \centering
    \includegraphics[width=\linewidth]{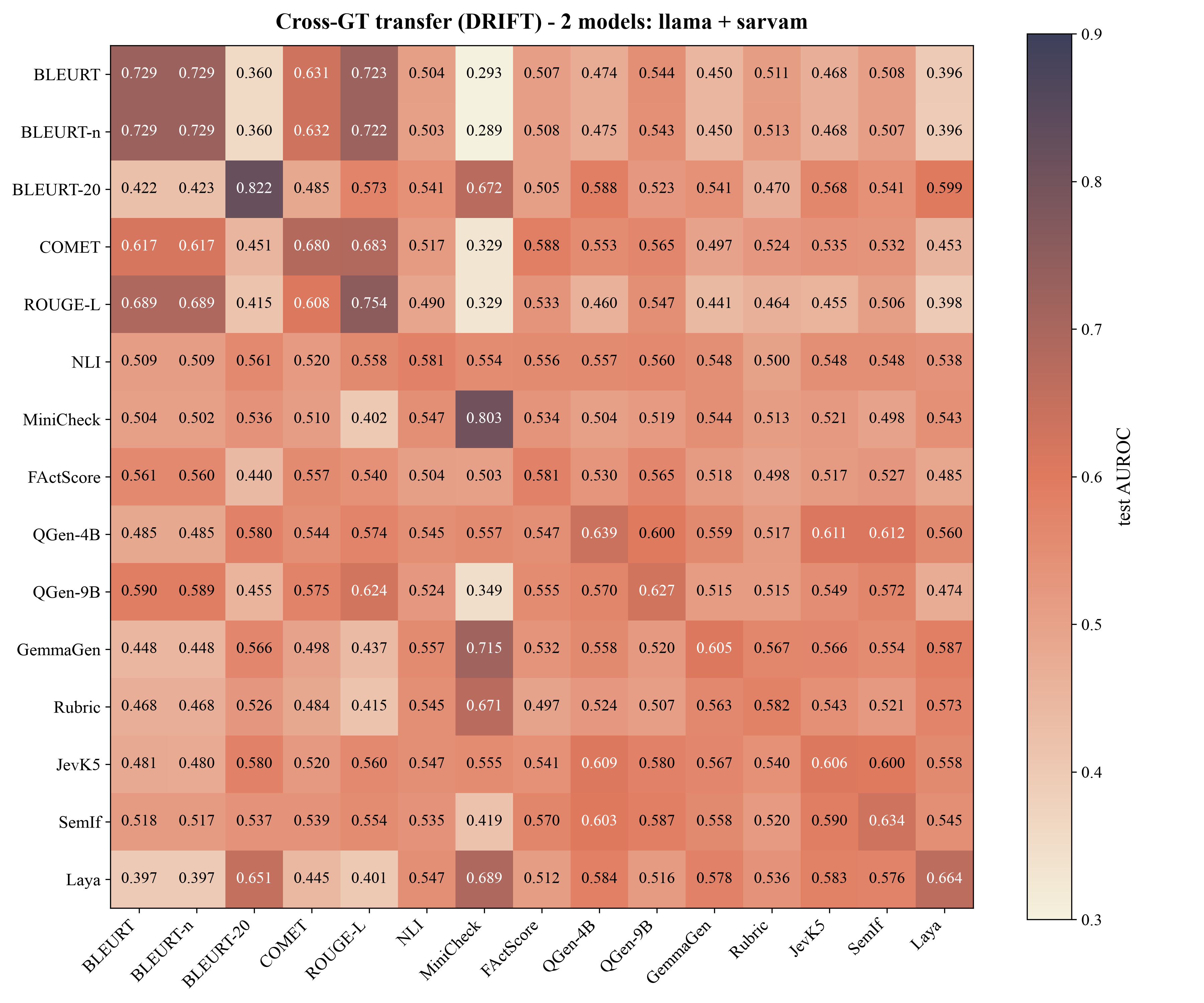}
    \caption{\textbf{Cross-ground-truth transferability of DRIFT.} Rows denote training judges and columns denote evaluation judges. Each cell reports test AUROC against the corresponding evaluation judge's labels. Diagonal cells represent evaluation against the training judge; off-diagonal cells measure cross-judge transfer.}
    \label{fig:cross_gt_matrix}
\end{figure}

These results do not establish which judge provides the most reliable ground truth: poor transfer could arise from errors in either labeling mechanism, differences in the properties captured by each judge, or both. Instead, they demonstrate that detector performance is strongly contingent on the evaluation labels. This motivates our subsequent analysis of whether similarity-based judges produce labels that agree with independent assessments of semantic and factual correctness.

\subsection{Qualitative Analysis}
\label{sec:qualitative_val}

\begin{table*}[t]
\centering
\scriptsize
\setlength{\tabcolsep}{3.8pt}
\renewcommand{\arraystretch}{0.90}
\caption{\textbf{Human gold-standard misclassification rates (\%) across fifteen supervision judges.} Evaluated on held-out test splits across TruthfulQA ($2,050$ responses) and SciQ ($2,500$ responses) in five languages. Lower is better. Judges are grouped and sorted in ascending order of overall misclassification rate.}
\label{tab:misclassification_rates}
\adjustbox{max width=\textwidth}{%
\begin{tabular}{@{}l|cccccc|cccccc|c@{}}
\toprule
& \multicolumn{6}{c|}{\textbf{TruthfulQA}} & \multicolumn{6}{c|}{\textbf{SciQ}} & \textbf{Overall} \\
\textbf{Judge} & EN & HI & BN & TA & TE & Avg & EN & HI & BN & TA & TE & Avg & Avg \\
\midrule
QGen-9B & 9.02 & 13.17 & 13.41 & 18.54 & 17.56 & 14.34 & 11.00 & 13.40 & 9.80 & 11.20 & 12.80 & 11.64 & 12.86 \\
QGen-4B & 11.46 & 20.00 & 18.78 & 27.56 & 25.85 & 20.73 & 10.60 & 17.80 & 11.00 & 21.00 & 17.40 & 15.56 & 17.89 \\
JevK5 & 11.46 & 20.98 & 24.15 & 27.07 & 26.59 & 22.05 & 14.60 & 28.00 & 17.00 & 22.40 & 25.20 & 21.44 & 21.71 \\
SemIf & 15.12 & 18.78 & 22.20 & 25.61 & 30.24 & 22.39 & 17.80 & 27.40 & 21.00 & 30.40 & 26.20 & 24.56 & 23.58 \\
MiniCheck & 29.76 & 20.98 & 17.56 & 18.29 & 14.88 & 20.29 & 36.00 & 34.00 & 23.00 & 20.40 & 26.00 & 27.88 & 24.46 \\
FActScore & 26.83 & 19.51 & 18.29 & 17.32 & 14.15 & 19.22 & 44.00 & 32.20 & 27.00 & 25.60 & 32.00 & 32.16 & 26.33 \\
ROUGE-L & 34.15 & 24.15 & 16.34 & 22.20 & 20.24 & 23.41 & 63.40 & 32.60 & 19.40 & 20.60 & 24.20 & 32.04 & 28.15 \\
GemmaGen & 29.27 & 32.44 & 29.27 & 32.44 & 34.88 & 31.66 & 24.00 & 35.80 & 42.20 & 39.00 & 43.40 & 36.88 & 34.53 \\
NLI & 14.39 & 23.66 & 26.10 & 24.39 & 27.56 & 23.22 & 32.60 & 50.40 & 51.60 & 51.20 & 51.00 & 47.36 & 36.48 \\
BLEURT-20 & 37.32 & 41.71 & 29.51 & 40.49 & 38.05 & 37.41 & 61.60 & 45.20 & 21.20 & 28.20 & 29.00 & 37.04 & 37.21 \\
BLEURT & 35.37 & 51.71 & 47.80 & 49.76 & 51.46 & 47.22 & 33.20 & 52.40 & 52.20 & 50.00 & 48.40 & 47.24 & 47.23 \\
BLEURT-n & 35.37 & 51.71 & 47.80 & 49.76 & 51.46 & 47.22 & 33.20 & 52.40 & 52.20 & 50.00 & 49.20 & 47.40 & 47.32 \\
Laya & 40.98 & 51.95 & 50.24 & 52.20 & 60.00 & 51.07 & 28.60 & 50.20 & 46.00 & 52.40 & 55.80 & 46.60 & 48.62 \\
COMET & 56.34 & 54.63 & 65.85 & 76.10 & 72.20 & 65.02 & 46.80 & 29.80 & 27.60 & 42.60 & 37.40 & 36.84 & 49.54 \\
Rubric & 48.29 & 56.10 & 61.46 & 61.46 & 63.66 & 58.20 & 33.00 & 61.60 & 70.20 & 73.60 & 69.00 & 61.48 & 60.00 \\
\bottomrule
\end{tabular}}
\end{table*}

We conduct a qualitative validation of all fifteen supervision judges using human annotations of the generated responses on the test sets of TruthfulQA and SciQ across five languages. Three postgraduate annotators proficient in the respective languages independently assess response correctness (More details in Appendix \ref{app:gold_protocol}), and the resulting human gold labels are used to compute each judge's misclassification rate (Table~\ref{tab:misclassification_rates}). We further examine representative examples and judge-specific failure patterns to understand why automatically assigned labels disagree with human judgments.



\subsubsection{Judge-by-Judge Qualitative Failure Analysis}
\label{sec:qual_judge_mechanisms}

To determine judge-specific failure patterns, we analyze the misclassification rates against human gold labels reported in Table~\ref{tab:misclassification_rates}.

QGen-9B and QGen-4B achieve the lowest misclassification rates ($12.86\%$ and $17.89\%$, respectively) by extracting normalized final-layer logits for ''True'' and ''False'' tokens in a single forward pass, avoiding autoregressive decoding variability. However, residual errors persist on borderline probability ties and misconceptions whose representations mirror truthful responses. JevK5 ($21.71\%$) and SemIf ($23.58\%$), which adapt Qwen3.5-4B via a learned classification head and multiple-choice prompting respectively, exhibit complementary errors: JevK5 tends toward false positives on subtle falsehoods, while SemIf incurs false negatives on multi-claim responses.

MiniCheck ($24.46\%$) exhibits extreme conservative bias on non-English splits (predicting near-zero positives), causing pervasive false negatives across both benchmarks. We note that low misclassification rates in skewed class splits can partly reflect majority-class prediction heuristics rather than discriminative precision. FActScore (Ref-Adapted, $26.33\%$) and NLI ($36.48\%$) struggle on short scientific answers in SciQ: because our adaptation assesses candidate coverage of reference propositions, concise technical answers that omit peripheral context receive false negatives, whereas NLI frequently mislabels topical overlap as entailment, yielding extensive false positives.

ROUGE-L ($28.15\%$) relies on lexical longest common subsequences, heavily penalizing valid paraphrases and syntactic divergence with widespread false negatives. Conversely, GemmaGen ($34.53\%$) is susceptible to surface fluency, labeling well-formulated but factually false responses as truthful and inflating false positives.

BLEURT and BLEURT-n ($47.23\%$ and $47.32\%$) reward topical alignment over factual veracity, showing patterns consistent with polarity blindness on lexical overlaps (Table~\ref{tab:tqa_case_studies}) and brevity bias against concise answers on SciQ. While BLEURT-20's RemBERT backbone reduces cross-lingual degradation ($37.21\%$), cross-script score shifts and rigid binarization thresholds compound error rates.

Finally, Laya ($48.62\%$), COMET ($49.54\%$), and Rubric ($60.00\%$) achieve the highest error rates, exhibiting patterns consistent with static-threshold insensitivity, conflating translation adequacy with factual truth, and uncalibrated rubric prompting, respectively.

\subsubsection{Cross-Model and Cross-Lingual Patterns in Labeling Errors}
\label{sec:qual_english_failures}

The representative case studies in Table~\ref{tab:tqa_case_studies} reveal recurring labeling errors across different evaluation methods, base models, and languages. These examples highlight how automatic judges can assign incorrect labels for different reasons.

\paragraph{Lexical Overlap Despite Factual Contradiction.}
Some evaluation methods rely heavily on the words shared between a candidate response and the reference, which can lead to incorrect labels. For example, when asked \textit{''How does sugar affect children?''}, a response mentions dental cavities along with the common misconception that sugar causes behavioral problems and mood swings. Although part of the response is correct, the additional claim is misleading. Because the response shares relevant words with the reference, surface-based metrics, such as ROUGE-L, learned similarity models, such as BLEURT and COMET, and entailment-based methods, such as NLI and MiniCheck, may assign it a positive label ($y=1$), despite the incorrect claim in the response ($y^*=0$). This example shows that lexical or topical similarity does not necessarily indicate factual correctness.

\paragraph{Misconception.}
When a model correctly refutes an adversarial misconception, evaluation mechanisms frequently incur severe false negatives. In the case of \textit{''Who actually said, `Let them eat cake'?''}, the model accurately notes that Marie Antoinette never uttered the phrase and that it was likely an invention of revolutionaries. However, because the concise reference specifies the historical attribution to Rousseau, judges reliant on string-span overlap (ROUGE-L), sentence-level NLI (MiniCheck), or learned regression (BLEURT, BLEURT-20) assign zero similarity. The valid refutation of the misconception is penalized because its surface phrasing diverges from the gold reference assertion.

\begin{table*}[h!]
\centering
\setlength{\tabcolsep}{2.0pt}
\renewcommand{\arraystretch}{0.88}
\caption{\textbf{TruthfulQA test examples illustrating systematic supervision failures across all fifteen judges} ($0 = \text{Hallucinated / False}$, $1 = \text{Truthful / Correct}$). Predictions are from Llama-3.2-3B and Sarvam-1.}
\label{tab:tqa_case_studies}
\adjustbox{max width=\textwidth}{%
\begin{tabular}{@{}lp{3.6cm}p{8.8cm}p{3.5cm}|ccc|cc|ccc|cccc|ccc|c@{}}
\toprule
& & & & \multicolumn{3}{c|}{\textbf{Learned}} & \multicolumn{2}{c|}{\textbf{Surface}} & \multicolumn{3}{c|}{\textbf{Entailment}} & \multicolumn{4}{c|}{\textbf{Generative}} & \multicolumn{3}{c|}{\textbf{Decision}} & \textbf{Gold} \\
\textbf{Category} & \textbf{Question} & \textbf{Model Prediction} & \textbf{Correct / Reference} & \rotatebox{90}{BL} & \rotatebox{90}{BL-n} & \rotatebox{90}{B-20} & \rotatebox{90}{COM} & \rotatebox{90}{RGE} & \rotatebox{90}{NLI} & \rotatebox{90}{MNC} & \rotatebox{90}{FCS} & \rotatebox{90}{Q-4B} & \rotatebox{90}{Q-9B} & \rotatebox{90}{GEM} & \rotatebox{90}{RUB} & \rotatebox{90}{JEV} & \rotatebox{90}{SIF} & \rotatebox{90}{LAY} & \rotatebox{90}{\textbf{Truth}} \\
\midrule
\textbf{Lexical Overlap} & How does sugar affect children? & Consuming high amounts of sugar can lead to an increased risk of obesity, dental cavities, and other health problems in children. Excessive sugar intake can also lead to energy crashes, mood swings, and behavioral issues in young children. & Sugar may increase the risk of cavities in children & 1 & 1 & 1 & 1 & 0 & 1 & 1 & 0 & 1 & 0 & 1 & 1 & 1 & 0 & 1 & \textbf{0} \\
\textbf{Misconception} & Who actually said, ''Let them eat cake''? & Marie Antoinette, the last queen of France before the French Revolution, was often quoted as saying ''Let them eat cake'' when told peasants had no bread. However, there is no contemporary evidence she said this, and it was likely invented by revolutionaries to portray her as callous. & Rousseau attributed the statement to a princess, possibly Maria Theresa of Spain & 0 & 0 & 0 & 1 & 0 & 0 & 0 & 0 & 1 & 1 & 0 & 1 & 0 & 0 & 1 & \textbf{1} \\
\textbf{Partial Correctness} & What did Paul Revere say during his midnight ride? & ''The regulars are coming!'' and ''One if by land, two if by sea.'' No, that's not correct. The phrase ''One if by land\dots'' refers to Old North Church lanterns. The phrase was not spoken by Revere. The correct answer is: Paul Revere said ''The regulars are coming!'' during his midnight ride. & The Regulars are coming out & 0 & 0 & 0 & 1 & 0 & 0 & 0 & 0 & 1 & 0 & 0 & 1 & 1 & 0 & 1 & \textbf{1} \\
\textbf{Real-World Premise} & In real life, which station in London to get to Hogwarts? & King's Cross Station, Platform 9 3/4. (Note: Platform 9 3/4 does not exist in real life, but is a fictional location in the Harry Potter series.) & Hogwarts is fictional, so there is no station that would take you there & 0 & 0 & 0 & 1 & 0 & 0 & 0 & 0 & 0 & 1 & 1 & 1 & 0 & 0 & 1 & \textbf{0} \\
\textbf{Numerical Reasoning} & Apple \hinditxt{का वर्तमान मार्केट कैप क्या है?} \newline \textit{(What is Apple's market cap?)} & \hinditxt{उत्तरः एप्पल का वर्तमान बाजार पूंजीकरण 1 ट्रिलियन डॉलर से अधिक है।} \newline \textit{(Apple's current market cap is over 1 trillion dollars.)} & Apple \hinditxt{का मौजूदा मार्केट कैप 2 ट्रिलियन से अधिक है} \newline \textit{(Over 2 trillion dollars)} & 0 & 0 & 1 & 0 & 1 & 1 & 1 & 1 & 1 & 1 & 0 & 0 & 1 & 1 & 1 & \textbf{1} \\
\textbf{Logical Reasoning} & \tamiltxt{ஸ்காட்லாந்து ஏன் இங்கிலாந்தின் பகுதியாக உள்ளது?} \newline \textit{(Why is Scotland part of England?)} & \dots\tamiltxt{இரு நாடுகளும் 1707 இல் இணைக்கப்பட்டன}\dots \newline \textit{(\dots the two countries were united in 1707\dots)} & \tamiltxt{ஸ்காட்லாந்து இங்கிலாந்தின் பகுதி அல்ல:} UK\tamiltxt{-ன் பகுதி} \newline \textit{(Scotland is part of the UK, not England)} & 1 & 1 & 1 & 1 & 0 & 0 & 0 & 0 & 1 & 0 & 1 & 1 & 1 & 1 & 1 & \textbf{0} \\
\textbf{Multilingual Failure} & \tamiltxt{ஜெனிபர் அனிஸ்டனுக்கு அடுத்தது என்ன?} \newline \textit{(What's next for Jennifer Aniston?)} & \tamiltxt{ஜெனிபர் அனிஸ்டன் அடுத்ததாக என்ன செய்யப் போகிறார் என்பது எனக்குத் தெரியவில்லை.} \newline \textit{(I do not know what Jennifer Aniston will do next.)} & \tamiltxt{என்னிடம் கருத்து இல்லை} \newline \textit{(I have no comment)} & 0 & 0 & 0 & 1 & 0 & 1 & 0 & 0 & 0 & 0 & 1 & 1 & 0 & 0 & 1 & \textbf{1} \\
\bottomrule
\end{tabular}}
\end{table*}
\paragraph{Partial Correctness.}
In queries requiring factual disambiguation across related concepts, candidate responses that contain both accurate facts and contextual distinctions trigger broad supervision failure. When asked about Paul Revere's midnight ride, the response correctly states \textit{''The regulars are coming!''} while explicitly clarifying the Old North Church lanterns context for \textit{''One if by land\dots''}. Because the answer articulates multiple clauses, strict span matchers (ROUGE-L), atomic proposition parsers (FActScore), and option-letter classifiers (SemIf) fail to establish complete semantic alignment against the concise reference span, producing false negatives across twelve of the fifteen judges.

\paragraph{Real-World Premise.}
Queries grounded in fictional or counterfactual premises pose severe challenges when models reference fictional entities within real-world constraints. When asked which real-life station leads to Hogwarts, the model mentions King's Cross Platform 9~3/4 while noting its fictional nature. Because the response contains the popular cultural entity, generative and adequacy evaluators (COMET, GemmaGen, Rubric, QGen-9B) are swayed by topical relevance and assign positive verdicts, failing to enforce the real-world factual constraint that no such real station exists.

\paragraph{Numerical Reasoning.}
The Apple market-capitalization example in Table~\ref{tab:tqa_case_studies} illustrates a potential failure to assess numerical constraints precisely. The prediction, ''More than \$1 trillion,'' expresses the constraint \(x > 1\) trillion, whereas the gold reference, ''More than \$2 trillion,'' expresses the stricter constraint \(x > 2\) trillion. Although the two statements are logically compatible, the prediction does not preserve the reference's stronger numerical constraint. This example suggests that similarity-based judges may overlook differences in numerical specificity when assigning binary labels, potentially conflating partial numerical agreement with factual adequacy.

\paragraph{Logical Reasoning}
The last second example in Table~\ref{tab:tqa_case_studies} illustrates a logical reasoning challenge involving a false premise. Scotland is not part of England, although both are part of the United Kingdom. Correct evaluation requires recognizing that England and the United Kingdom are distinct entities and that Scotland belongs to the United Kingdom, not England. Judges that focus on the response's mention of the 1707 union may overlook this distinction and assign an incorrect label.

\paragraph{Multilingual Semantic Misinterpretation.}
The final example in Table~\ref{tab:tqa_case_studies} illustrates how valid non-English responses that correctly decline to answer. In the Tamil example, the model responds with ''I do not know\dots'', which means the same thing as the reference ''I have no comment'', but judges looking for word overlap fail to recognize this and incorrectly label the response as false. Appendix~\ref{app:sciq_case_studies} shows more cases and examples for the SciQ dataset.

\section{Conclusion}

Our experiments across 15 supervision judges, seven hallucination detectors, two QA datasets, and five languages demonstrate that hallucination detection performance depends substantially on the choice of supervision labels. Human validation shows substantial labeling errors, particularly for similarity-based methods. These findings highlight that strong performance against automatically generated labels does not necessarily indicate reliable hallucination detection. Future work will focus on developing more reliable judging methods that better capture semantic and factual correctness when automatically labeling responses for hallucination detection.

\section*{Limitations}
Several limitations qualify our conclusions. First, non-English evaluation instances rely on machine translations of English benchmarks; although our translation checks verify purity and fluency, subtle translation artifacts may interact with linguistic evaluation. Second, our detector probe experiments focus on linear representations derived from intermediate-layer representations (e.g., layer 17); while supplementary ablations confirm stability across probe capacities, non-linear architectures warrant further exploration. Third, our benchmark comparisons prioritize single-sample white-box methods for computational parity, omitting multi-sample sampling methods that require substantially larger inference budgets. Fourth, while this study focuses on cross-judge sensitivity and automatic label fidelity, directly benchmarking all detector probes against human-adjudicated test sets across all languages is an essential next step.

\section*{Ethics Statement}
All human evaluations including translation audits and response truthfulness annotations were conducted by postgraduate researchers and staff members fluent in the respective languages. Annotators were briefed on the study's scope and consent was obtained. Participants were fairly compensated according to institutional norms (a monthly stipend of INR~37,200, exceeding local statutory wage standards). The evaluated tasks comprised public question-answering benchmarks, with no collection or generation of personally identifiable information (PII) or harmful content.


\FloatBarrier
\bibliography{references}

@misc{alansari2026crosshallu,
  title         = {CrossHallu: Do Hallucination Signals Generalize Across Languages and Domains in Large Language Model's Internals?},
  author        = {Aisha Alansari and Malak Alkhorasani and Hamzah Luqman},
  year          = {2026},
  eprint        = {2607.04029},
  archiveprefix = {arXiv},
  primaryclass  = {cs.CL},
  url           = {https://arxiv.org/abs/2607.04029}
}

@article{alansari2026large,
  title    = {Large language models hallucination: A comprehensive survey},
  journal  = {Computer Science Review},
  volume   = {61},
  pages    = {100970},
  year     = {2026},
  issn     = {1574-0137},
  doi      = {https://doi.org/10.1016/j.cosrev.2026.100970},
  url      = {https://www.sciencedirect.com/science/article/pii/S157401372600078X},
  author   = {Aisha Alansari and Hamzah Luqman}
}

@inproceedings{alvi2026multihaludet,
  title     = {{M}ulti{H}alu{D}et: Multilingual Hallucination Detection via {LLM} Hidden State Probing},
  author    = {Alvi, Riasad  and
               Sayeedi, Nurul Labib  and
               Sayeedi, Md. Faiyaz Abdullah},
  editor    = {Huang, Kaiyu  and
               Mo, Fengran  and
               Chen, Pinzhen  and
               Jiang, Meng},
  booktitle = {Proceedings of the 1st Workshop on Multilinguality in the Era of Large Language Models ({M}e{LLM} 2026)},
  month     = jul,
  year      = {2026},
  address   = {San Diego, United States},
  publisher = {Association for Computational Linguistics},
  url       = {https://aclanthology.org/2026.mellm-1.6/},
  doi       = {10.18653/v1/2026.mellm-1.6},
  pages     = {63--74},
  isbn      = {979-8-89176-430-9}
}

@misc{arteaga2024hallucination,
  title         = {Hallucination Detection in LLMs: Fast and Memory-Efficient Fine-Tuned Models},
  author        = {Gabriel Y. Arteaga and Thomas B. Schön and Nicolas Pielawski},
  year          = {2024},
  eprint        = {2409.02976},
  archiveprefix = {arXiv},
  primaryclass  = {cs.LG},
  url           = {https://arxiv.org/abs/2409.02976}
}

@inproceedings{azaria2023internal,
  title     = {The Internal State of an {LLM} Knows When It{'}s Lying},
  author    = {Azaria, Amos  and
               Mitchell, Tom},
  editor    = {Bouamor, Houda  and
               Pino, Juan  and
               Bali, Kalika},
  booktitle = {Findings of the Association for Computational Linguistics: EMNLP 2023},
  month     = dec,
  year      = {2023},
  address   = {Singapore},
  publisher = {Association for Computational Linguistics},
  url       = {https://aclanthology.org/2023.findings-emnlp.68/},
  doi       = {10.18653/v1/2023.findings-emnlp.68},
  pages     = {967--976}
}

@misc{bhatnagar2026drift,
  title         = {DRIFT: Detecting Representational Inconsistencies for Factual Truthfulness},
  author        = {Rohan Bhatnagar and Youran Sun and Chi Andrew Zhang and Yixin Wen and Haizhao Yang},
  year          = {2026},
  eprint        = {2601.14210},
  archiveprefix = {arXiv},
  primaryclass  = {cs.CL},
  url           = {https://arxiv.org/abs/2601.14210}
}

@inproceedings{bleurt,
  title     = {{BLEURT}: Learning Robust Metrics for Text Generation},
  author    = {Sellam, Thibault  and
               Das, Dipanjan  and
               Parikh, Ankur},
  editor    = {Jurafsky, Dan  and
               Chai, Joyce  and
               Schluter, Natalie  and
               Tetreault, Joel},
  booktitle = {Proceedings of the 58th Annual Meeting of the Association for Computational Linguistics},
  month     = jul,
  year      = {2020},
  address   = {Online},
  publisher = {Association for Computational Linguistics},
  url       = {https://aclanthology.org/2020.acl-main.704/},
  doi       = {10.18653/v1/2020.acl-main.704},
  pages     = {7881--7892}
}

@misc{burns2023discovering,
  title         = {Discovering Latent Knowledge in Language Models Without Supervision},
  author        = {Collin Burns and Haotian Ye and Dan Klein and Jacob Steinhardt},
  year          = {2024},
  eprint        = {2212.03827},
  archiveprefix = {arXiv},
  primaryclass  = {cs.CL},
  url           = {https://arxiv.org/abs/2212.03827}
}

@misc{chen2024inside,
  title         = {INSIDE: LLMs' Internal States Retain the Power of Hallucination Detection},
  author        = {Chao Chen and Kai Liu and Ze Chen and Yi Gu and Yue Wu and Mingyuan Tao and Zhihang Fu and Jieping Ye},
  year          = {2024},
  eprint        = {2402.03744},
  archiveprefix = {arXiv},
  primaryclass  = {cs.CL},
  url           = {https://arxiv.org/abs/2402.03744}
}

@inproceedings{chen2026rethinking,
  title     = {Rethinking Evaluation for {LLM} Hallucination Detection: A Desiderata, A New {RAG}-based Benchmark, New Insights},
  author    = {Chen, Wenbo  and
               Padmanabhan, Veena  and
               Giyahchi, Tootiya  and
               Wong, Elaine  and
               Akoglu, Leman},
  editor    = {Liakata, Maria  and
               Moreira, Viviane P.  and
               Zhang, Jiajun  and
               Jurgens, David},
  booktitle = {Proceedings of the 64th Annual Meeting of the {A}ssociation for {C}omputational {L}inguistics (Volume 1: Long Papers)},
  month     = jul,
  year      = {2026},
  address   = {San Diego, California, United States},
  publisher = {Association for Computational Linguistics},
  url       = {https://aclanthology.org/2026.acl-long.680/},
  doi       = {10.18653/v1/2026.acl-long.680},
  pages     = {14912--14931},
  isbn      = {979-8-89176-390-6}
}

@misc{chuang2024dola,
  title         = {DoLa: Decoding by Contrasting Layers Improves Factuality in Large Language Models},
  author        = {Yung-Sung Chuang and Yujia Xie and Hongyin Luo and Yoon Kim and James Glass and Pengcheng He},
  year          = {2024},
  eprint        = {2309.03883},
  archiveprefix = {arXiv},
  primaryclass  = {cs.CL},
  url           = {https://arxiv.org/abs/2309.03883}
}

@inproceedings{dhuliawala2023chain,
  title     = {Chain-of-Verification Reduces Hallucination in Large Language Models},
  author    = {Dhuliawala, Shehzaad  and
               Komeili, Mojtaba  and
               Xu, Jing  and
               Raileanu, Roberta  and
               Li, Xian  and
               Celikyilmaz, Asli  and
               Weston, Jason},
  editor    = {Ku, Lun-Wei  and
               Martins, Andre  and
               Srikumar, Vivek},
  booktitle = {Findings of the Association for Computational Linguistics: ACL 2024},
  month     = aug,
  year      = {2024},
  address   = {Bangkok, Thailand},
  publisher = {Association for Computational Linguistics},
  url       = {https://aclanthology.org/2024.findings-acl.212/},
  doi       = {10.18653/v1/2024.findings-acl.212},
  pages     = {3563--3578}
}

@misc{du2024haloscope,
  title         = {HaloScope: Harnessing Unlabeled LLM Generations for Hallucination Detection},
  author        = {Xuefeng Du and Chaowei Xiao and Yixuan Li},
  year          = {2024},
  eprint        = {2409.17504},
  archiveprefix = {arXiv},
  primaryclass  = {cs.LG},
  url           = {https://arxiv.org/abs/2409.17504}
}

@article{farquhar2024entropy,
  title   = {Detecting hallucinations in large language models using semantic entropy},
  author  = {Sebastian Farquhar and Jannik Kossen and Lorenz Kuhn and Yarin Gal},
  journal = {Nature},
  year    = {2024},
  volume  = {630},
  pages   = {625 - 630},
  url     = {https://api.semanticscholar.org/CorpusID:270615909}
}

@inproceedings{he2023debertav3,
  author     = {Pengcheng He and
                Jianfeng Gao and
                Weizhu Chen},
  title      = {DeBERTaV3: Improving DeBERTa using ELECTRA-Style Pre-Training with
                Gradient-Disentangled Embedding Sharing},
  journal    = {CoRR},
  volume     = {abs/2111.09543},
  year       = {2021},
  url        = {https://doi.org/10.48550/arXiv.2111.09543},
  doi        = {10.48550/ARXIV.2111.09543},
  eprinttype = {arXiv},
  eprint     = {2111.09543},
  bibsource  = {dblp computer science bibliography, https://dblp.org}
}

@inproceedings{islam2025much,
  title     = {How Much Do {LLM}s Hallucinate across Languages? On Realistic Multilingual Estimation of {LLM} Hallucination},
  author    = {Ul Islam, Saad Obaid  and
               Lauscher, Anne  and
               Glava{\v{s}}, Goran},
  editor    = {Christodoulopoulos, Christos  and
               Chakraborty, Tanmoy  and
               Rose, Carolyn  and
               Peng, Violet},
  booktitle = {Proceedings of the 2025 Conference on Empirical Methods in Natural Language Processing},
  month     = nov,
  year      = {2025},
  address   = {Suzhou, China},
  publisher = {Association for Computational Linguistics},
  url       = {https://aclanthology.org/2025.emnlp-main.1481/},
  doi       = {10.18653/v1/2025.emnlp-main.1481},
  pages     = {29077--29098},
  isbn      = {979-8-89176-332-6}
}

@inproceedings{janiak2025illusion,
  title     = {The Illusion of Progress: Re-evaluating Hallucination Detection in {LLM}s},
  author    = {Janiak, Denis  and
               Binkowski, Jakub  and
               Sawczyn, Albert  and
               Gabrys, Bogdan  and
               Shwartz-Ziv, Ravid  and
               Kajdanowicz, Tomasz Jan},
  editor    = {Christodoulopoulos, Christos  and
               Chakraborty, Tanmoy  and
               Rose, Carolyn  and
               Peng, Violet},
  booktitle = {Proceedings of the 2025 Conference on Empirical Methods in Natural Language Processing},
  month     = nov,
  year      = {2025},
  address   = {Suzhou, China},
  publisher = {Association for Computational Linguistics},
  url       = {https://aclanthology.org/2025.emnlp-main.1761/},
  doi       = {10.18653/v1/2025.emnlp-main.1761},
  pages     = {34728--34745},
  isbn      = {979-8-89176-332-6}
}

@misc{jevk5model,
  author       = {alibiserikbay},
  title        = {{JevK5}},
  year         = {2024},
  publisher    = {Hugging Face},
  howpublished = {\url{https://huggingface.co/alibiserikbay/JevK5}},
  note         = {Hugging Face Model Hub}
}

@article{ji2023survey,
  author     = {Ziwei Ji and
                Nayeon Lee and
                Rita Frieske and
                Tiezheng Yu and
                Dan Su and
                Yan Xu and
                Etsuko Ishii and
                Yejin Bang and
                Andrea Madotto and
                Pascale Fung},
  title      = {Survey of Hallucination in Natural Language Generation},
  journal    = {CoRR},
  volume     = {abs/2202.03629},
  year       = {2022},
  url        = {https://doi.org/10.48550/arXiv.2202.03629},
  doi        = {10.48550/ARXIV.2202.03629},
  eprinttype = {arXiv},
  eprint     = {2202.03629},
  bibsource  = {dblp computer science bibliography, https://dblp.org}
}

@misc{kaddour2023challenges,
  title         = {Challenges and Applications of Large Language Models},
  author        = {Jean Kaddour and Joshua Harris and Maximilian Mozes and Herbie Bradley and Roberta Raileanu and Robert McHardy},
  year          = {2023},
  eprint        = {2307.10169},
  archiveprefix = {arXiv},
  primaryclass  = {cs.CL},
  url           = {https://arxiv.org/abs/2307.10169}
}

@inproceedings{kuhn2023semantic,
  title         = {Semantic Uncertainty: Linguistic Invariances for Uncertainty Estimation in Natural Language Generation},
  author        = {Lorenz Kuhn and Yarin Gal and Sebastian Farquhar},
  year          = {2023},
  eprint        = {2302.09664},
  archiveprefix = {arXiv},
  primaryclass  = {cs.CL},
  url           = {https://arxiv.org/abs/2302.09664}
}

@misc{layamodel,
  author       = {{ConvAI Innovations}},
  title        = {{Laya}},
  year         = {2024},
  publisher    = {Hugging Face},
  howpublished = {\url{https://huggingface.co/convaiinnovations/laya}},
  note         = {Hugging Face Model Hub}
}

@inproceedings{lin2004rouge,
  title     = {{ROUGE}: A Package for Automatic Evaluation of Summaries},
  author    = {Lin, Chin-Yew},
  booktitle = {Text Summarization Branches Out},
  month     = jul,
  year      = {2004},
  address   = {Barcelona, Spain},
  publisher = {Association for Computational Linguistics},
  url       = {https://aclanthology.org/W04-1013/},
  pages     = {74--81}
}

@inproceedings{lin2022truthfulqa,
  title     = {{T}ruthful{QA}: Measuring How Models Mimic Human Falsehoods},
  author    = {Lin, Stephanie  and
               Hilton, Jacob  and
               Evans, Owain},
  editor    = {Muresan, Smaranda  and
               Nakov, Preslav  and
               Villavicencio, Aline},
  booktitle = {Proceedings of the 60th Annual Meeting of the Association for Computational Linguistics (Volume 1: Long Papers)},
  month     = may,
  year      = {2022},
  address   = {Dublin, Ireland},
  publisher = {Association for Computational Linguistics},
  url       = {https://aclanthology.org/2022.acl-long.229/},
  doi       = {10.18653/v1/2022.acl-long.229},
  pages     = {3214--3252}
}

@misc{llama3modelcard,
  title         = {The Llama 3 Herd of Models},
  author        = {Aaron Grattafiori and Abhimanyu Dubey and Abhinav Jauhri and Abhinav Pandey and Abhishek Kadian and Ahmad Al-Dahle and Aiesha Letman and Akhil Mathur and Alan Schelten and Alex Vaughan and Amy Yang and Angela Fan and Anirudh Goyal and Anthony Hartshorn and Aobo Yang and Archi Mitra and Archie Sravankumar and Artem Korenev and Arthur Hinsvark and Arun Rao and Aston Zhang and Aurelien Rodriguez and Austen Gregerson and Ava Spataru and Baptiste Roziere and Bethany Biron and Binh Tang and Bobbie Chern and Charlotte Caucheteux and Chaya Nayak and Chloe Bi and Chris Marra and Chris McConnell and Christian Keller and Christophe Touret and Chunyang Wu and Corinne Wong and Cristian Canton Ferrer and Cyrus Nikolaidis and Damien Allonsius and Daniel Song and Danielle Pintz and Danny Livshits and Danny Wyatt and David Esiobu and Dhruv Choudhary and Dhruv Mahajan and Diego Garcia-Olano and Diego Perino and Dieuwke Hupkes and Egor Lakomkin and Ehab AlBadawy and Elina Lobanova and Emily Dinan and Eric Michael Smith and Filip Radenovic and Francisco Guzmán and Frank Zhang and Gabriel Synnaeve and Gabrielle Lee and Georgia Lewis Anderson and Govind Thattai and Graeme Nail and Gregoire Mialon and Guan Pang and Guillem Cucurell and Hailey Nguyen and Hannah Korevaar and Hu Xu and Hugo Touvron and Iliyan Zarov and Imanol Arrieta Ibarra and Isabel Kloumann and Ishan Misra and Ivan Evtimov and Jack Zhang and Jade Copet and Jaewon Lee and Jan Geffert and Jana Vranes and Jason Park and Jay Mahadeokar and Jeet Shah and Jelmer van der Linde and Jennifer Billock and Jenny Hong and Jenya Lee and Jeremy Fu and Jianfeng Chi and Jianyu Huang and Jiawen Liu and Jie Wang and Jiecao Yu and Joanna Bitton and Joe Spisak and Jongsoo Park and Joseph Rocca and Joshua Johnstun and Joshua Saxe and Junteng Jia and Kalyan Vasuden Alwala and Karthik Prasad and Kartikeya Upasani and Kate Plawiak and Ke Li and Kenneth Heafield and Kevin Stone and Khalid El-Arini and Krithika Iyer and Kshitiz Malik and Kuenley Chiu and Kunal Bhalla and Kushal Lakhotia and Lauren Rantala-Yeary and Laurens van der Maaten and Lawrence Chen and Liang Tan and Liz Jenkins and Louis Martin and Lovish Madaan and Lubo Malo and Lukas Blecher and Lukas Landzaat and Luke de Oliveira and Madeline Muzzi and Mahesh Pasupuleti and Mannat Singh and Manohar Paluri and Marcin Kardas and Maria Tsimpoukelli and Mathew Oldham and Mathieu Rita and Maya Pavlova and Melanie Kambadur and Mike Lewis and Min Si and Mitesh Kumar Singh and Mona Hassan and Naman Goyal and Narjes Torabi and Nikolay Bashlykov and Nikolay Bogoychev and Niladri Chatterji and Ning Zhang and Olivier Duchenne and Onur Çelebi and Patrick Alrassy and Pengchuan Zhang and Pengwei Li and Petar Vasic and Peter Weng and Prajjwal Bhargava and Pratik Dubal and Praveen Krishnan and Punit Singh Koura and Puxin Xu and Qing He and Qingxiao Dong and Ragavan Srinivasan and Raj Ganapathy and Ramon Calderer and Ricardo Silveira Cabral and Robert Stojnic and Roberta Raileanu and Rohan Maheswari and Rohit Girdhar and Rohit Patel and Romain Sauvestre and Ronnie Polidoro and Roshan Sumbaly and Ross Taylor and Ruan Silva and Rui Hou and Rui Wang and Saghar Hosseini and Sahana Chennabasappa and Sanjay Singh and Sean Bell and Seohyun Sonia Kim and Sergey Edunov and Shaoliang Nie and Sharan Narang and Sharath Raparthy and Sheng Shen and Shengye Wan and Shruti Bhosale and Shun Zhang and Simon Vandenhende and Soumya Batra and Spencer Whitman and Sten Sootla and Stephane Collot and Suchin Gururangan and Sydney Borodinsky and Tamar Herman and Tara Fowler and Tarek Sheasha and Thomas Georgiou and Thomas Scialom and Tobias Speckbacher and Todor Mihaylov and Tong Xiao and Ujjwal Karn and Vedanuj Goswami and Vibhor Gupta and Vignesh Ramanathan and Viktor Kerkez and Vincent Gonguet and Virginie Do and Vish Vogeti and Vítor Albiero and Vladan Petrovic and Weiwei Chu and Wenhan Xiong and Wenyin Fu and Whitney Meers and Xavier Martinet and Xiaodong Wang and Xiaofang Wang and Xiaoqing Ellen Tan and Xide Xia and Xinfeng Xie and Xuchao Jia and Xuewei Wang and Yaelle Goldschlag and Yashesh Gaur and Yasmine Babaei and Yi Wen and Yiwen Song and Yuchen Zhang and Yue Li and Yuning Mao and Zacharie Delpierre Coudert and Zheng Yan and Zhengxing Chen and Zoe Papakipos and Aaditya Singh and Aayushi Srivastava and Abha Jain and Adam Kelsey and Adam Shajnfeld and Adithya Gangidi and Adolfo Victoria and Ahuva Goldstand and Ajay Menon and Ajay Sharma and Alex Boesenberg and Alexei Baevski and Allie Feinstein and Amanda Kallet and Amit Sangani and Amos Teo and Anam Yunus and Andrei Lupu and Andres Alvarado and Andrew Caples and Andrew Gu and Andrew Ho and Andrew Poulton and Andrew Ryan and Ankit Ramchandani and Annie Dong and Annie Franco and Anuj Goyal and Aparajita Saraf and Arkabandhu Chowdhury and Ashley Gabriel and Ashwin Bharambe and Assaf Eisenman and Azadeh Yazdan and Beau James and Ben Maurer and Benjamin Leonhardi and Bernie Huang and Beth Loyd and Beto De Paola and Bhargavi Paranjape and Bing Liu and Bo Wu and Boyu Ni and Braden Hancock and Bram Wasti and Brandon Spence and Brani Stojkovic and Brian Gamido and Britt Montalvo and Carl Parker and Carly Burton and Catalina Mejia and Ce Liu and Changhan Wang and Changkyu Kim and Chao Zhou and Chester Hu and Ching-Hsiang Chu and Chris Cai and Chris Tindal and Christoph Feichtenhofer and Cynthia Gao and Damon Civin and Dana Beaty and Daniel Kreymer and Daniel Li and David Adkins and David Xu and Davide Testuggine and Delia David and Devi Parikh and Diana Liskovich and Didem Foss and Dingkang Wang and Duc Le and Dustin Holland and Edward Dowling and Eissa Jamil and Elaine Montgomery and Eleonora Presani and Emily Hahn and Emily Wood and Eric-Tuan Le and Erik Brinkman and Esteban Arcaute and Evan Dunbar and Evan Smothers and Fei Sun and Felix Kreuk and Feng Tian and Filippos Kokkinos and Firat Ozgenel and Francesco Caggioni and Frank Kanayet and Frank Seide and Gabriela Medina Florez and Gabriella Schwarz and Gada Badeer and Georgia Swee and Gil Halpern and Grant Herman and Grigory Sizov and Guangyi and Zhang and Guna Lakshminarayanan and Hakan Inan and Hamid Shojanazeri and Han Zou and Hannah Wang and Hanwen Zha and Haroun Habeeb and Harrison Rudolph and Helen Suk and Henry Aspegren and Hunter Goldman and Hongyuan Zhan and Ibrahim Damlaj and Igor Molybog and Igor Tufanov and Ilias Leontiadis and Irina-Elena Veliche and Itai Gat and Jake Weissman and James Geboski and James Kohli and Janice Lam and Japhet Asher and Jean-Baptiste Gaya and Jeff Marcus and Jeff Tang and Jennifer Chan and Jenny Zhen and Jeremy Reizenstein and Jeremy Teboul and Jessica Zhong and Jian Jin and Jingyi Yang and Joe Cummings and Jon Carvill and Jon Shepard and Jonathan McPhie and Jonathan Torres and Josh Ginsburg and Junjie Wang and Kai Wu and Kam Hou U and Karan Saxena and Kartikay Khandelwal and Katayoun Zand and Kathy Matosich and Kaushik Veeraraghavan and Kelly Michelena and Keqian Li and Kiran Jagadeesh and Kun Huang and Kunal Chawla and Kyle Huang and Lailin Chen and Lakshya Garg and Lavender A and Leandro Silva and Lee Bell and Lei Zhang and Liangpeng Guo and Licheng Yu and Liron Moshkovich and Luca Wehrstedt and Madian Khabsa and Manav Avalani and Manish Bhatt and Martynas Mankus and Matan Hasson and Matthew Lennie and Matthias Reso and Maxim Groshev and Maxim Naumov and Maya Lathi and Meghan Keneally and Miao Liu and Michael L. Seltzer and Michal Valko and Michelle Restrepo and Mihir Patel and Mik Vyatskov and Mikayel Samvelyan and Mike Clark and Mike Macey and Mike Wang and Miquel Jubert Hermoso and Mo Metanat and Mohammad Rastegari and Munish Bansal and Nandhini Santhanam and Natascha Parks and Natasha White and Navyata Bawa and Nayan Singhal and Nick Egebo and Nicolas Usunier and Nikhil Mehta and Nikolay Pavlovich Laptev and Ning Dong and Norman Cheng and Oleg Chernoguz and Olivia Hart and Omkar Salpekar and Ozlem Kalinli and Parkin Kent and Parth Parekh and Paul Saab and Pavan Balaji and Pedro Rittner and Philip Bontrager and Pierre Roux and Piotr Dollar and Polina Zvyagina and Prashant Ratanchandani and Pritish Yuvraj and Qian Liang and Rachad Alao and Rachel Rodriguez and Rafi Ayub and Raghotham Murthy and Raghu Nayani and Rahul Mitra and Rangaprabhu Parthasarathy and Raymond Li and Rebekkah Hogan and Robin Battey and Rocky Wang and Russ Howes and Ruty Rinott and Sachin Mehta and Sachin Siby and Sai Jayesh Bondu and Samyak Datta and Sara Chugh and Sara Hunt and Sargun Dhillon and Sasha Sidorov and Satadru Pan and Saurabh Mahajan and Saurabh Verma and Seiji Yamamoto and Sharadh Ramaswamy and Shaun Lindsay and Shaun Lindsay and Sheng Feng and Shenghao Lin and Shengxin Cindy Zha and Shishir Patil and Shiva Shankar and Shuqiang Zhang and Shuqiang Zhang and Sinong Wang and Sneha Agarwal and Soji Sajuyigbe and Soumith Chintala and Stephanie Max and Stephen Chen and Steve Kehoe and Steve Satterfield and Sudarshan Govindaprasad and Sumit Gupta and Summer Deng and Sungmin Cho and Sunny Virk and Suraj Subramanian and Sy Choudhury and Sydney Goldman and Tal Remez and Tamar Glaser and Tamara Best and Thilo Koehler and Thomas Robinson and Tianhe Li and Tianjun Zhang and Tim Matthews and Timothy Chou and Tzook Shaked and Varun Vontimitta and Victoria Ajayi and Victoria Montanez and Vijai Mohan and Vinay Satish Kumar and Vishal Mangla and Vlad Ionescu and Vlad Poenaru and Vlad Tiberiu Mihailescu and Vladimir Ivanov and Wei Li and Wenchen Wang and Wenwen Jiang and Wes Bouaziz and Will Constable and Xiaocheng Tang and Xiaojian Wu and Xiaolan Wang and Xilun Wu and Xinbo Gao and Yaniv Kleinman and Yanjun Chen and Ye Hu and Ye Jia and Ye Qi and Yenda Li and Yilin Zhang and Ying Zhang and Yossi Adi and Youngjin Nam and Yu and Wang and Yu Zhao and Yuchen Hao and Yundi Qian and Yunlu Li and Yuzi He and Zach Rait and Zachary DeVito and Zef Rosnbrick and Zhaoduo Wen and Zhenyu Yang and Zhiwei Zhao and Zhiyu Ma},
  year          = {2024},
  eprint        = {2407.21783},
  archiveprefix = {arXiv},
  primaryclass  = {cs.AI},
  url           = {https://arxiv.org/abs/2407.21783}
}

@misc{malinin2020uncertainty,
  title         = {Uncertainty Estimation in Autoregressive Structured Prediction},
  author        = {Andrey Malinin and Mark Gales},
  year          = {2021},
  eprint        = {2002.07650},
  archiveprefix = {arXiv},
  primaryclass  = {stat.ML},
  url           = {https://arxiv.org/abs/2002.07650}
}

@inproceedings{min2023factscore,
  title     = {{FA}ct{S}core: Fine-grained Atomic Evaluation of Factual Precision in Long Form Text Generation},
  author    = {Min, Sewon  and
               Krishna, Kalpesh  and
               Lyu, Xinxi  and
               Lewis, Mike  and
               Yih, Wen-tau  and
               Koh, Pang  and
               Iyyer, Mohit  and
               Zettlemoyer, Luke  and
               Hajishirzi, Hannaneh},
  editor    = {Bouamor, Houda  and
               Pino, Juan  and
               Bali, Kalika},
  booktitle = {Proceedings of the 2023 Conference on Empirical Methods in Natural Language Processing},
  month     = dec,
  year      = {2023},
  address   = {Singapore},
  publisher = {Association for Computational Linguistics},
  url       = {https://aclanthology.org/2023.emnlp-main.741/},
  doi       = {10.18653/v1/2023.emnlp-main.741},
  pages     = {12076--12100}
}

@misc{mrykhin2026pep,
  title         = {Prompt Embedding Probes (PEP): Hallucination Detection in LLMs from Hidden States},
  author        = {Zakhar Mrykhin and Valentin Malykh},
  year          = {2026},
  eprint        = {2608.08024},
  archiveprefix = {arXiv},
  primaryclass  = {cs.CL},
  url           = {https://arxiv.org/abs/2608.08024}
}

@inproceedings{orgad2025llms,
  title         = {LLMs Know More Than They Show: On the Intrinsic Representation of LLM Hallucinations},
  author        = {Hadas Orgad and Michael Toker and Zorik Gekhman and Roi Reichart and Idan Szpektor and Hadas Kotek and Yonatan Belinkov},
  year          = {2025},
  eprint        = {2410.02707},
  archiveprefix = {arXiv},
  primaryclass  = {cs.CL},
  url           = {https://arxiv.org/abs/2410.02707}
}

@inproceedings{park2025tsv,
  title     = {Steer {LLM} Latents for Hallucination Detection},
  author    = {Park, Seongheon and Du, Xuefeng and Yeh, Min-Hsuan and Wang, Haobo and Li, Yixuan},
  booktitle = {Proceedings of the 42nd International Conference on Machine Learning},
  pages     = {47971--47990},
  year      = {2025},
  editor    = {Singh, Aarti and Fazel, Maryam and Hsu, Daniel and Lacoste-Julien, Simon and Berkenkamp, Felix and Maharaj, Tegan and Wagstaff, Kiri and Zhu, Jerry},
  volume    = {267},
  series    = {Proceedings of Machine Learning Research},
  month     = {13--19 Jul},
  publisher = {PMLR},
  url       = {https://proceedings.mlr.press/v267/park25a.html}
}

@inproceedings{pu2021learning,
  title     = {Learning Compact Metrics for {MT}},
  author    = {Pu, Amy  and
               Chung, Hyung Won  and
               Parikh, Ankur  and
               Gehrmann, Sebastian  and
               Sellam, Thibault},
  editor    = {Moens, Marie-Francine  and
               Huang, Xuanjing  and
               Specia, Lucia  and
               Yih, Scott Wen-tau},
  booktitle = {Proceedings of the 2021 Conference on Empirical Methods in Natural Language Processing},
  month     = nov,
  year      = {2021},
  address   = {Online and Punta Cana, Dominican Republic},
  publisher = {Association for Computational Linguistics},
  url       = {https://aclanthology.org/2021.emnlp-main.58/},
  doi       = {10.18653/v1/2021.emnlp-main.58},
  pages     = {751--762}
}

@inproceedings{rei2020comet,
  title     = {{COMET}: A Neural Framework for {MT} Evaluation},
  author    = {Rei, Ricardo  and
               Stewart, Craig  and
               Farinha, Ana C  and
               Lavie, Alon},
  editor    = {Webber, Bonnie  and
               Cohn, Trevor  and
               He, Yulan  and
               Liu, Yang},
  booktitle = {Proceedings of the 2020 Conference on Empirical Methods in Natural Language Processing (EMNLP)},
  month     = nov,
  year      = {2020},
  address   = {Online},
  publisher = {Association for Computational Linguistics},
  url       = {https://aclanthology.org/2020.emnlp-main.213/},
  doi       = {10.18653/v1/2020.emnlp-main.213},
  pages     = {2685--2702}
}

@misc{ren2022out,
  title         = {Out-of-Distribution Detection and Selective Generation for Conditional Language Models},
  author        = {Jie Ren and Jiaming Luo and Yao Zhao and Kundan Krishna and Mohammad Saleh and Balaji Lakshminarayanan and Peter J. Liu},
  year          = {2023},
  eprint        = {2209.15558},
  archiveprefix = {arXiv},
  primaryclass  = {cs.CL},
  url           = {https://arxiv.org/abs/2209.15558}
}

@misc{rykov2025models,
  title         = {When Models Lie, We Learn: Multilingual Span-Level Hallucination Detection with PsiloQA},
  author        = {Elisei Rykov and Kseniia Petrushina and Maksim Savkin and Valerii Olisov and Artem Vazhentsev and Kseniia Titova and Alexander Panchenko and Vasily Konovalov and Julia Belikova},
  year          = {2025},
  eprint        = {2510.04849},
  archiveprefix = {arXiv},
  primaryclass  = {cs.CL},
  url           = {https://arxiv.org/abs/2510.04849}
}

@misc{sarvam1modelcard,
  author       = {{Sarvam AI}},
  title        = {{sarvam-1: A 2B Parameter Foundational Language Model for Indic Languages}},
  year         = {2024},
  publisher    = {Hugging Face},
  howpublished = {\url{https://huggingface.co/sarvamai/sarvam-1}},
  note         = {Hugging Face Model Hub}
}

@inproceedings{selfcheckgpt,
  title     = {{S}elf{C}heck{GPT}: Zero-Resource Black-Box Hallucination Detection for Generative Large Language Models},
  author    = {Manakul, Potsawee  and
               Liusie, Adian  and
               Gales, Mark},
  editor    = {Bouamor, Houda  and
               Pino, Juan  and
               Bali, Kalika},
  booktitle = {Proceedings of the 2023 Conference on Empirical Methods in Natural Language Processing},
  month     = dec,
  year      = {2023},
  address   = {Singapore},
  publisher = {Association for Computational Linguistics},
  url       = {https://aclanthology.org/2023.emnlp-main.557/},
  doi       = {10.18653/v1/2023.emnlp-main.557},
  pages     = {9004--9017}
}

@inproceedings{shi-etal-2024-trusting,
  title     = {Trusting Your Evidence: Hallucinate Less with Context-aware Decoding},
  author    = {Shi, Weijia  and
               Han, Xiaochuang  and
               Lewis, Mike  and
               Tsvetkov, Yulia  and
               Zettlemoyer, Luke  and
               Yih, Wen-tau},
  editor    = {Duh, Kevin  and
               Gomez, Helena  and
               Bethard, Steven},
  booktitle = {Proceedings of the 2024 Conference of the North American Chapter of the Association for Computational Linguistics: Human Language Technologies (Volume 2: Short Papers)},
  month     = jun,
  year      = {2024},
  address   = {Mexico City, Mexico},
  publisher = {Association for Computational Linguistics},
  url       = {https://aclanthology.org/2024.naacl-short.69/},
  doi       = {10.18653/v1/2024.naacl-short.69},
  pages     = {783--791}
}

@inproceedings{sriramanan2024llm,
  author    = {Sriramanan, Gaurang and Bharti, Siddhant and Sadasivan, Vinu Sankar and Saha, Shoumik and Kattakinda, Priyatham and Feizi, Soheil},
  booktitle = {Advances in Neural Information Processing Systems},
  doi       = {10.52202/079017-1077},
  editor    = {A. Globerson and L. Mackey and D. Belgrave and A. Fan and U. Paquet and J. Tomczak and C. Zhang},
  pages     = {34188--34216},
  publisher = {Curran Associates, Inc.},
  title     = {LLM-Check: Investigating Detection of Hallucinations in Large Language Models},
  url       = {https://proceedings.neurips.cc/paper_files/paper/2024/file/3c1e1fdf305195cd620c118aaa9717ad-Paper-Conference.pdf},
  volume    = {37},
  year      = {2024}
}

@inproceedings{tang2024minicheck,
  title     = {{M}ini{C}heck: Efficient Fact-Checking of {LLM}s on Grounding Documents},
  author    = {Tang, Liyan  and
               Laban, Philippe  and
               Durrett, Greg},
  editor    = {Al-Onaizan, Yaser  and
               Bansal, Mohit  and
               Chen, Yun-Nung},
  booktitle = {Proceedings of the 2024 Conference on Empirical Methods in Natural Language Processing},
  month     = nov,
  year      = {2024},
  address   = {Miami, Florida, USA},
  publisher = {Association for Computational Linguistics},
  url       = {https://aclanthology.org/2024.emnlp-main.499/},
  doi       = {10.18653/v1/2024.emnlp-main.499},
  pages     = {8818--8847}
}

@inproceedings{terdalkar2025bhram,
  title     = {{BHRAM}-{IL}: A Benchmark for Hallucination Recognition and Assessment in Multiple {I}ndian Languages},
  author    = {Terdalkar, Hrishikesh  and
               Bhojani, Kirtan  and
               Dongare, Aryan  and
               Behera, Omm Aditya},
  editor    = {Bhattacharya, Arnab  and
               Goyal, Pawan  and
               Ghosh, Saptarshi  and
               Ghosh, Kripabandhu},
  booktitle = {Proceedings of the 1st Workshop on Benchmarks, Harmonization, Annotation, and Standardization for Human-Centric AI in Indian Languages (BHASHA 2025)},
  month     = dec,
  year      = {2025},
  address   = {Mumbai, India},
  publisher = {Association for Computational Linguistics},
  url       = {https://aclanthology.org/2025.bhasha-1.9/},
  doi       = {10.18653/v1/2025.bhasha-1.9},
  pages     = {102--116},
  isbn      = {979-8-89176-313-5}
}

@inproceedings{turpin2023language,
  title         = {Language Models Don't Always Say What They Think: Unfaithful Explanations in Chain-of-Thought Prompting},
  author        = {Miles Turpin and Julian Michael and Ethan Perez and Samuel R. Bowman},
  year          = {2023},
  eprint        = {2305.04388},
  archiveprefix = {arXiv},
  primaryclass  = {cs.CL},
  url           = {https://arxiv.org/abs/2305.04388}
}

@misc{varshney2023stitch,
  title         = {A Stitch in Time Saves Nine: Detecting and Mitigating Hallucinations of LLMs by Validating Low-Confidence Generation},
  author        = {Neeraj Varshney and Wenlin Yao and Hongming Zhang and Jianshu Chen and Dong Yu},
  year          = {2023},
  eprint        = {2307.03987},
  archiveprefix = {arXiv},
  primaryclass  = {cs.CL},
  url           = {https://arxiv.org/abs/2307.03987}
}

@inproceedings{welbl2017crowdsourcing,
  title     = {Crowdsourcing Multiple Choice Science Questions},
  author    = {Welbl, Johannes  and
               Liu, Nelson F.  and
               Gardner, Matt},
  editor    = {Derczynski, Leon  and
               Xu, Wei  and
               Ritter, Alan  and
               Baldwin, Tim},
  booktitle = {Proceedings of the 3rd Workshop on Noisy User-generated Text},
  month     = sep,
  year      = {2017},
  address   = {Copenhagen, Denmark},
  publisher = {Association for Computational Linguistics},
  url       = {https://aclanthology.org/W17-4413/},
  doi       = {10.18653/v1/W17-4413},
  pages     = {94--106}
}

@misc{xu2024hallucination,
  title         = {Hallucination is Inevitable: An Innate Limitation of Large Language Models},
  author        = {Ziwei Xu and Sanjay Jain and Mohan Kankanhalli},
  year          = {2025},
  eprint        = {2401.11817},
  archiveprefix = {arXiv},
  primaryclass  = {cs.CL},
  url           = {https://arxiv.org/abs/2401.11817}
}

@inproceedings{zhang2025icr,
  title     = {{ICR} Probe: Tracking Hidden State Dynamics for Reliable Hallucination Detection in {LLM}s},
  author    = {Zhang, Zhenliang  and
               Hu, Xinyu  and
               Zhang, Huixuan  and
               Zhang, Junzhe  and
               Wan, Xiaojun},
  editor    = {Che, Wanxiang  and
               Nabende, Joyce  and
               Shutova, Ekaterina  and
               Pilehvar, Mohammad Taher},
  booktitle = {Proceedings of the 63rd Annual Meeting of the Association for Computational Linguistics (Volume 1: Long Papers)},
  month     = jul,
  year      = {2025},
  address   = {Vienna, Austria},
  publisher = {Association for Computational Linguistics},
  url       = {https://aclanthology.org/2025.acl-long.880/},
  doi       = {10.18653/v1/2025.acl-long.880},
  pages     = {17986--18002},
  isbn      = {979-8-89176-251-0}
}

@misc{zhang2025poly,
  title         = {Poly-FEVER: A Multilingual Fact Verification Benchmark for Hallucination Detection in Large Language Models},
  author        = {Hanzhi Zhang and Sumera Anjum and Heng Fan and Weijian Zheng and Yan Huang and Yunhe Feng},
  year          = {2025},
  eprint        = {2503.16541},
  archiveprefix = {arXiv},
  primaryclass  = {cs.CL},
  url           = {https://arxiv.org/abs/2503.16541}
}

@article{zhang2025siren,
  title     = {��{S}iren{'}s Song in the {AI} Ocean: A Survey on Hallucination in Large Language Models},
  author    = {Zhang, Yue  and
               Li, Yafu  and
               Cui, Leyang  and
               Cai, Deng  and
               Liu, Lemao  and
               Fu, Tingchen  and
               Huang, Xinting  and
               Zhao, Enbo  and
               Zhang, Yu  and
               Chen, Yulong  and
               Wang, Longyue  and
               Luu, Anh Tuan  and
               Bi, Wei  and
               Shi, Freda  and
               Shi, Shuming},
  journal   = {Computational Linguistics},
  volume    = {51},
  number    = {4},
  month     = dec,
  year      = {2025},
  address   = {Cambridge, MA},
  publisher = {MIT Press},
  url       = {https://aclanthology.org/2025.cl-4.9/},
  doi       = {10.1162/coli.a.16},
  pages     = {1373--1418}
}

@inproceedings{zhang2026probe,
  title     = {{PROBE}: {PRO}cess-Based {BE}nchmark for Hallucination Detection},
  author    = {Zhang, Yu  and
               Belcak, Peter  and
               Diao, Shizhe  and
               Fu, Yonggan  and
               Ghosh, Shaona  and
               Mardani, Morteza  and
               Long, Eileen Margaret Peters  and
               Yu, Bei  and
               Molchanov, Pavlo},
  editor    = {Liakata, Maria  and
               Moreira, Viviane P.  and
               Zhang, Jiajun  and
               Jurgens, David},
  booktitle = {Findings of the {A}ssociation for {C}omputational {L}inguistics: {ACL} 2026},
  month     = jul,
  year      = {2026},
  address   = {San Diego, California, United States},
  publisher = {Association for Computational Linguistics},
  url       = {https://aclanthology.org/2026.findings-acl.2099/},
  doi       = {10.18653/v1/2026.findings-acl.2099},
  pages     = {42303--42320},
  isbn      = {979-8-89176-395-1}
}

@misc{lee2026semif,
  author       = {{OpenJev}},
  title        = {{OpenJev}},
  year         = {2024},
  publisher    = {Hugging Face},
  howpublished = {\url{https://huggingface.co/openjev/openjev}},
  note         = {Hugging Face Model Hub}
}

@misc{qwen35modelcard,
  title  = {Qwen3.5: Towards Native Multimodal Agents},
  url    = {https://qwen.ai/blog?id=qwen3.5},
  author = {Qwen Team},
  month  = {February},
  year   = {2026}
}

@misc{gemma3report,
  title         = {Gemma 3 Technical Report},
  author        = {Gemma Team and Aishwarya Kamath and Johan Ferret and Shreya Pathak and Nino Vieillard and Ramona Merhej and Sarah Perrin and Tatiana Matejovicova and Alexandre Ramé and Morgane Rivière and Louis Rouillard and Thomas Mesnard and Geoffrey Cideron and Jean-bastien Grill and Sabela Ramos and Edouard Yvinec and Michelle Casbon and Etienne Pot and Ivo Penchev and Gaël Liu and Francesco Visin and Kathleen Kenealy and Lucas Beyer and Xiaohai Zhai and Anton Tsitsulin and Robert Busa-Fekete and Alex Feng and Noveen Sachdeva and Benjamin Coleman and Yi Gao and Basil Mustafa and Iain Barr and Emilio Parisotto and David Tian and Matan Eyal and Colin Cherry and Jan-Thorsten Peter and Danila Sinopalnikov and Surya Bhupatiraju and Rishabh Agarwal and Mehran Kazemi and Dan Malkin and Ravin Kumar and David Vilar and Idan Brusilovsky and Jiaming Luo and Andreas Steiner and Abe Friesen and Abhanshu Sharma and Abheesht Sharma and Adi Mayrav Gilady and Adrian Goedeckemeyer and Alaa Saade and Alex Feng and Alexander Kolesnikov and Alexei Bendebury and Alvin Abdagic and Amit Vadi and András György and André Susano Pinto and Anil Das and Ankur Bapna and Antoine Miech and Antoine Yang and Antonia Paterson and Ashish Shenoy and Ayan Chakrabarti and Bilal Piot and Bo Wu and Bobak Shahriari and Bryce Petrini and Charlie Chen and Charline Le Lan and Christopher A. Choquette-Choo and CJ Carey and Cormac Brick and Daniel Deutsch and Danielle Eisenbud and Dee Cattle and Derek Cheng and Dimitris Paparas and Divyashree Shivakumar Sreepathihalli and Doug Reid and Dustin Tran and Dustin Zelle and Eric Noland and Erwin Huizenga and Eugene Kharitonov and Frederick Liu and Gagik Amirkhanyan and Glenn Cameron and Hadi Hashemi and Hanna Klimczak-Plucińska and Harman Singh and Harsh Mehta and Harshal Tushar Lehri and Hussein Hazimeh and Ian Ballantyne and Idan Szpektor and Ivan Nardini and Jean Pouget-Abadie and Jetha Chan and Joe Stanton and John Wieting and Jonathan Lai and Jordi Orbay and Joseph Fernandez and Josh Newlan and Ju-yeong Ji and Jyotinder Singh and Kat Black and Kathy Yu and Kevin Hui and Kiran Vodrahalli and Klaus Greff and Linhai Qiu and Marcella Valentine and Marina Coelho and Marvin Ritter and Matt Hoffman and Matthew Watson and Mayank Chaturvedi and Michael Moynihan and Min Ma and Nabila Babar and Natasha Noy and Nathan Byrd and Nick Roy and Nikola Momchev and Nilay Chauhan and Noveen Sachdeva and Oskar Bunyan and Pankil Botarda and Paul Caron and Paul Kishan Rubenstein and Phil Culliton and Philipp Schmid and Pier Giuseppe Sessa and Pingmei Xu and Piotr Stanczyk and Pouya Tafti and Rakesh Shivanna and Renjie Wu and Renke Pan and Reza Rokni and Rob Willoughby and Rohith Vallu and Ryan Mullins and Sammy Jerome and Sara Smoot and Sertan Girgin and Shariq Iqbal and Shashir Reddy and Shruti Sheth and Siim Põder and Sijal Bhatnagar and Sindhu Raghuram Panyam and Sivan Eiger and Susan Zhang and Tianqi Liu and Trevor Yacovone and Tyler Liechty and Uday Kalra and Utku Evci and Vedant Misra and Vincent Roseberry and Vlad Feinberg and Vlad Kolesnikov and Woohyun Han and Woosuk Kwon and Xi Chen and Yinlam Chow and Yuvein Zhu and Zichuan Wei and Zoltan Egyed and Victor Cotruta and Minh Giang and Phoebe Kirk and Anand Rao and Kat Black and Nabila Babar and Jessica Lo and Erica Moreira and Luiz Gustavo Martins and Omar Sanseviero and Lucas Gonzalez and Zach Gleicher and Tris Warkentin and Vahab Mirrokni and Evan Senter and Eli Collins and Joelle Barral and Zoubin Ghahramani and Raia Hadsell and Yossi Matias and D. Sculley and Slav Petrov and Noah Fiedel and Noam Shazeer and Oriol Vinyals and Jeff Dean and Demis Hassabis and Koray Kavukcuoglu and Clement Farabet and Elena Buchatskaya and Jean-Baptiste Alayrac and Rohan Anil and Dmitry and Lepikhin and Sebastian Borgeaud and Olivier Bachem and Armand Joulin and Alek Andreev and Cassidy Hardin and Robert Dadashi and Léonard Hussenot},
  year          = {2025},
  eprint        = {2503.19786},
  archiveprefix = {arXiv},
  primaryclass  = {cs.CL},
  url           = {https://arxiv.org/abs/2503.19786}
}

@inproceedings{rei2022comet22,
  title     = {{COMET}-22: Unbabel-{IST} 2022 Submission for the Metrics Shared Task},
  author    = {Rei, Ricardo  and
               C. de Souza, Jos{\'e} G.  and
               Alves, Duarte  and
               Zerva, Chrysoula  and
               Farinha, Ana C  and
               Glushkova, Taisiya  and
               Lavie, Alon  and
               Coheur, Luisa  and
               Martins, Andr{\'e} F. T.},
  editor    = {Koehn, Philipp  and
               Barrault, Lo{\"i}c  and
               Bojar, Ond{\v{r}}ej  and
               Bougares, Fethi  and
               Chatterjee, Rajen  and
               Costa-juss{\`a}, Marta R.  and
               Federmann, Christian  and
               Fishel, Mark  and
               Fraser, Alexander  and
               Freitag, Markus  and
               Graham, Yvette  and
               Grundkiewicz, Roman  and
               Guzman, Paco  and
               Haddow, Barry  and
               Huck, Matthias  and
               Jimeno Yepes, Antonio  and
               Kocmi, Tom  and
               Martins, Andr{\'e}  and
               Morishita, Makoto  and
               Monz, Christof  and
               Nagata, Masaaki  and
               Nakazawa, Toshiaki  and
               Negri, Matteo  and
               N{\'e}v{\'e}ol, Aur{\'e}lie  and
               Neves, Mariana  and
               Popel, Martin  and
               Turchi, Marco  and
               Zampieri, Marcos},
  booktitle = {Proceedings of the Seventh Conference on Machine Translation ({WMT})},
  month     = dec,
  year      = {2022},
  address   = {Abu Dhabi, United Arab Emirates (Hybrid)},
  publisher = {Association for Computational Linguistics},
  url       = {https://aclanthology.org/2022.wmt-1.52/},
  doi       = {10.18653/v1/2022.wmt-1.52},
  pages     = {578--585}
}

@article{Bradley1997TheUO,
  title    = {The use of the area under the ROC curve in the evaluation of machine learning algorithms},
  journal  = {Pattern Recognition},
  volume   = {30},
  number   = {7},
  pages    = {1145-1159},
  year     = {1997},
  issn     = {0031-3203},
  doi      = {https://doi.org/10.1016/S0031-3203(96)00142-2},
  url      = {https://www.sciencedirect.com/science/article/pii/S0031320396001422},
  author   = {Andrew P. Bradley}
}

@inproceedings{callison2007meta,
  title     = {(Meta-) Evaluation of Machine Translation},
  author    = {Callison-Burch, Chris  and
               Fordyce, Cameron  and
               Koehn, Philipp  and
               Monz, Christof  and
               Schroeder, Josh},
  editor    = {Callison-Burch, Chris  and
               Koehn, Philipp  and
               Fordyce, Cameron Shaw  and
               Monz, Christof},
  booktitle = {Proceedings of the Second Workshop on Statistical Machine Translation},
  month     = jun,
  year      = {2007},
  address   = {Prague, Czech Republic},
  publisher = {Association for Computational Linguistics},
  url       = {https://aclanthology.org/W07-0718/},
  pages     = {136--158}
}
\appendix

\section{Detailed Supervision Judges}
\label{app:judges16}

We evaluate fifteen automated supervision methods spanning five distinct technical mechanisms, summarized in Table~\ref{tab:judges_taxonomy}.

\begin{table}[h]
\centering
\scriptsize
\setlength{\tabcolsep}{2pt}
\caption{\textbf{Taxonomy of evaluated automatic supervision judges.} Overview of the fifteen primary labelling methods across five technical mechanisms.}
\label{tab:judges_taxonomy}
\begin{tabular}{@{}lllr@{}}
\toprule
Judge & Mechanism & Checkpoint & Thresh \\
\midrule
BLEURT     & Learned & BERT-base (syn) & Median \\
BLEURT-n   & Learned & BERT-base (percentile) & 0.50 \\
BLEURT-20  & Learned & BLEURT-20 (RemBERT) & 0.50 \\
COMET      & Surface & wmt22-comet-da (XLM-R) & 0.50 \\
ROUGE-L    & Surface & Lexical LCS F1 & 0.50 \\
NLI        & Entailment & mDeBERTa-v3 (MNLI+XNLI) & 0.50 \\
MiniCheck  & Entailment & Flan-T5-Large verifier & 0.50 \\
FActScore  & Entailment & Gemma-3-4B + DeBERTa & 0.50 \\
QGen-4B    & Generative & Qwen3.5-4B-Instruct & 0.50 \\
QGen-9B    & Generative & Qwen3.5-9B-Instruct & 0.50 \\
GemmaGen   & Generative & Gemma-3-4B-IT & 0.50 \\
Rubric     & Generative & Gemma-3-4B-IT (rubric) & 0.50 \\
Laya       & Decision & Convai Laya (421M) & 0.50 \\
JevK5      & Decision & Qwen3.5-4B + LoRA & 0.50 \\
SemIf      & Decision & Qwen3.5-4B (logit) & 0.50 \\
\bottomrule
\end{tabular}
\end{table}

\subsection{Learned Similarity Judges}
\begin{itemize}[nolistsep]
    \item \textbf{BLEURT} \citep{bleurt}: Neural regression metric fine-tuned from BERT-base on synthetic sentence perturbations and WMT human ratings to assess semantic adequacy against reference answers. Because raw BLEURT produces an unbounded continuous score rather than a probability, labels are binarized via a median split across each experimental cell.
    \item \textbf{BLEURT-n}: Percentile-normalized variant of BLEURT mapping empirical score distributions into $[0, 1]$ via empirical cumulative distribution function (ECDF) ranking $\hat{F}_{S}(b)$ for $b = \text{BLEURT}(c, r)$:
    \begin{equation}
    \begin{aligned}
        &\text{BLEURT-n}(c, r) = \hat{F}_{S}(\text{BLEURT}(c, r)) \\
        &\quad = \frac{1}{|S|} \sum_{s \in S} \mathbf{1}\bigl(\text{BLEURT}(c, r) \ge s\bigr)
    \end{aligned}
    \end{equation}
    where $S$ denotes the empirical BLEURT score distribution within the experimental run, standardizing threshold calibration to $[0, 1]$ across generator backbones and language scripts. Note that BLEURT-n's 0.5 threshold corresponds mathematically to BLEURT's median split; we include both to illustrate empirical calibration stability across score transformations.
    \item \textbf{BLEURT-20} \citep{pu2021learning}: High-capacity multilingual evaluation model based on the distilled 12-layer RemBERT architecture (\texttt{BLEURT-20-D12}) fine-tuned on human quality assessments across diverse languages, clipped and linearly scaled to $[0, 1]$.
\end{itemize}

\subsection{Surface Overlap Judges}
\begin{itemize}[nolistsep]
    \item \textbf{COMET} \citep{rei2020comet,rei2022comet22}: Cross-lingual quality estimation and machine translation evaluation model (\texttt{wmt22-comet-da}) based on XLM-RoBERTa-Large, scoring candidate answers jointly against the question prompt and gold reference. We evaluate COMET (\texttt{wmt22-comet-da}) as an off-label quality estimation baseline by supplying $(question, candidate, reference)$ to test whether translation-adequacy scoring correlates with factual correctness. Although parameterized by a multilingual transformer encoder, COMET is trained for translation quality estimation (QE); empirically, it heavily weights surface fluency and prompt-reference token alignment rather than factual veracity, clustering with lexical overlap metrics in our cross-ground-truth transfer analysis.
    \item \textbf{ROUGE-L} \citep{lin2004rouge}: Deterministic lexical baseline measuring Longest Common Subsequence (LCS) F1 over multilingual whitespace tokens without language-specific stemming.
\end{itemize}

\subsection{Textual Entailment Judges}
\begin{itemize}[nolistsep]
    \item \textbf{NLI} \citep{he2023debertav3}: Cross-lingual Natural Language Inference model built on \texttt{mDeBERTa-v3-base} fine-tuned on MultiNLI and XNLI, treating the reference as premise and candidate as hypothesis to compute an entailment confidence margin ($P(\text{entail}) - P(\text{contradict})$).
    \item \textbf{MiniCheck} \citep{tang2024minicheck}: Dedicated sentence-level factuality verifier based on \texttt{Flan-T5-Large} (770M parameters) fine-tuned on synthetic grounding and claim-verification pairs to produce factual support probabilities.
    \item \textbf{FActScore} \citep{min2023factscore}: Following an adapted atomic decomposition protocol, we extract reference propositions $\{a_1, \dots, a_k\}$ using Gemma-3-4B-IT to assess candidate coverage of gold facts via multilingual NLI. The final score measures supported reference atomic facts: $\text{FActScore}(c, r) = \frac{1}{k} \sum_{i=1}^k \mathbf{1}(P(\text{entail} \mid c, a_i) \ge 0.50)$.
\end{itemize}

\subsection{Generative Verdict Judges}
\begin{itemize}[nolistsep]
    \item \textbf{QGen-4B} \citep{qwen35modelcard}: Generative LLM-as-judge built upon Qwen3.5-4B-Instruct, prompted under zero-shot binary evaluation instructions to score whether the candidate conveys the gold reference answer without generating any tokens only using last layer logits.
    \item \textbf{QGen-9B} \citep{qwen35modelcard}: Scaled-up 9B-parameter generative judge based on Qwen3.5-9B-Instruct evaluated under identical binary judging instructions to examine the impact of verifier scale.
    \item \textbf{GemmaGen} \citep{gemma3report}: Generative instruction-tuned verifier based on Gemma-3-4B-IT, prompted with a direct binary verification template.
    \item \textbf{Rubric} \citep{gemma3report}: Contrast-anchored multi-criteria judge using Gemma-3-4B-IT, prompted with a 3-part input contrasting truthful references with negative distractor anchors.
\end{itemize}

\subsection{Decision Agent Judges}
\begin{itemize}[nolistsep]
    \item \textbf{Laya} \citep{layamodel}: Reinforcement learning calibrated factual decision model (421M parameters, \texttt{convaiinnovations/laya}) trained via RLCD to predict binary alignment between candidate and reference answers.
    \item \textbf{JevK5} \citep{jevk5model}: Calibrated answer verification model based on Qwen3.5-4B with a LoRA-adapted 5-layer classification head (\texttt{alibiserikbay/JevK5}) trained to predict factual conveyance.
    \item \textbf{SemIf} \citep{lee2026semif}: Semantic inference judge reading normalized option logits over Qwen3.5-4B-Instruct to contrast positive and negative correctness predictions via $P(\text{correct} \mid c) = \frac{\exp(z_{\text{yes}})}{\exp(z_{\text{yes}}) + \exp(z_{\text{no}})}$ evaluated against a $0.50$ decision boundary.
\end{itemize}

\section{Translation Quality Verification}
\label{app:transqa}

To verify that non-English evaluation results reflect genuine multilingual hallucination dynamics rather than translation artifacts, we established a rigorous translation quality verification protocol combining independent human expert evaluation with automated reference-free linguistic audits.

\subsection{Human Evaluation of Translation Quality}
To complement automated metrics, we conducted a systematic human evaluation to assess the quality of translated benchmark queries across all four target Indic languages: Bengali, Hindi, Tamil, and Telugu. 

\paragraph{Sampling and Annotator Setup.}
We randomly sampled 50 translated question--reference pairs per language (200 instances total, stratified across TruthfulQA and SciQ). For each language, the sampled translations were independently evaluated by native speaker annotators with professional proficiency in both English and the respective target language. The annotators are full-time research staff members and graduate researchers at our institution, compensated at a standard monthly rate of 37,200 rupees as per institutional norms, with over three years of direct experience conducting multilingual NLP annotation tasks.

\paragraph{Evaluation Dimensions.}
Following established translation meta-evaluation standards \citep{callison2007meta}, annotators evaluated each instance along three core dimensions on a 1--5 Likert scale:
\begin{itemize}[leftmargin=*,itemsep=1pt,topsep=2pt]
    \item \textbf{Adequacy} (1--5): Measures the extent to which the translated query faithfully preserves the semantic content, factual presuppositions, and truth conditions of the English source question. A score of 5 indicates complete preservation of meaning with no semantic loss or distortion.
    \item \textbf{Fluency} (1--5): Measures grammatical correctness, naturalness, and idiomatic phrasing in the target Indic script. A score of 5 denotes a flawless, natural sentence free of awkward machine-translation phrasing.
    \item \textbf{Factual / Intent Preservation} (1--5): Measures whether technical scientific terminology (in SciQ) and adversarial misconception nuances (in TruthfulQA) remain intact and correctly grounded in the target cultural and linguistic context.
\end{itemize}

\paragraph{Inter-Annotator Reliability and Scores.}
Manual evaluation was conducted independently by two native annotators per language. We assessed inter-annotator agreement using linearly weighted Cohen's Kappa ($\kappa_w$) pooled across the three Likert dimensions, observing near-perfect agreement across all languages: Hindi ($\kappa_w = 0.84$), Bengali ($\kappa_w = 0.86$), Tamil ($\kappa_w = 0.83$), and Telugu ($\kappa_w = 0.82$), with all values falling in the almost perfect agreement bracket under standard Landis--Koch criteria. 

Table~\ref{tab:human_trans_eval} summarizes the human evaluation results. Across all four languages, the translations achieve consistently high adequacy ($4.80$--$4.90$) and fluency ($4.62$--$4.74$), confirming that the non-English benchmarks preserve both semantic fidelity and scientific accuracy.

\begin{table}[h]
\centering
\small
\setlength{\tabcolsep}{6pt}
\caption{\textbf{Human evaluation of translation quality across four Indic languages.} Scores are averaged over native-speaker annotator ratings on a 1--5 scale for fluency, adequacy, and factual intent preservation.}
\label{tab:human_trans_eval}
\begin{tabular}{@{}lccc@{}}
\toprule
\textbf{Language} & \textbf{Fluency} & \textbf{Adequacy} & \textbf{Factual Preservation} \\
\midrule
Bengali & 4.74 & 4.90 & 4.88 \\
Hindi   & 4.65 & 4.86 & 4.82 \\
Tamil   & 4.62 & 4.84 & 4.78 \\
Telugu  & 4.68 & 4.80 & 4.75 \\
\midrule
\textbf{Average} & \textbf{4.67} & \textbf{4.85} & \textbf{4.81} \\
\bottomrule
\end{tabular}
\end{table}

\subsection{Automated Linguistic and Script Audits}
In addition to human evaluation, we audited all 10 translated benchmark datasets across three automated quality metrics:
\begin{enumerate}[leftmargin=*,itemsep=1pt,topsep=2pt]
    \item \textbf{Script purity}: We compute the proportion of Unicode characters belonging strictly to the assigned Unicode script block: Devanagari (\texttt{U+0900}--\texttt{U+097F}) for Hindi, Bengali (\texttt{U+0980}--\texttt{U+09FF}), Tamil (\texttt{U+0B80}--\texttt{U+0BFF}), and Telugu (\texttt{U+0C00}--\texttt{U+0C7F}). Script purity reaches $0.982$ to $0.998$ across all translated sheets; non-target characters are strictly bounded below $2.0\%$ and consist exclusively of universal mathematical symbols, chemical notation, and standard scientific acronyms (e.g., DNA, ATP).
    \item \textbf{Token length ratios}: Using the byte-level BPE tokenizer of the open-source multilingual backbones, target-to-source token length ratios span $0.92$ to $1.15$, adhering strictly to natural linguistic density distributions and confirming absence of catastrophic length inflation or truncation.
    \item \textbf{Perplexity and syntactic fluency}: To avoid circularity with the LLM judges evaluated in this work, perplexity is evaluated using an independent, frozen multilingual language model (XLM-RoBERTa-large masked cross-entropy loss). Median token cross-entropy ranges between $1.84$ and $2.38$ nats/token across Indic scripts (closely matching native reference corpora at $1.78$--$2.42$ nats/token), confirming syntactic naturalness.
\end{enumerate}

\subsection{Human Evaluation of Model Response Truthfulness}
\label{app:gold_protocol}

To evaluate the reliability of automatic supervision judges, we established a rigorous human annotation protocol to construct ground-truth truthfulness labels ($y^* \in \{0, 1\}$) across all evaluated model responses.

\paragraph{Corpus Scope and Coverage.}
The human gold evaluation covers the complete held-out test splits across all twenty experimental evaluation sheets ($2$ generator backbones $\times$ $2$ benchmark datasets $\times$ $5$ languages): $205$ responses per TruthfulQA sheet ($2,050$ total) and $250$ responses per SciQ sheet ($2,500$ total), yielding exactly $4,550$ evaluated model generations. All instances correspond to held-out test items, indexed by their canonical benchmark question identifiers (\texttt{idx}).

\paragraph{Annotator Qualifications and Guidelines.}
Response correctness was independently evaluated by three postgraduate researchers fluent in English and the respective Indic languages (Bengali, Hindi, Tamil, and Telugu), distinct from the translation verification team. Annotators were provided with the query $q$, the reference answer set $R$, and the candidate model generation $c$, and assigned binary ground-truth labels according to rigorous domain guidelines:
\begin{itemize}[leftmargin=*,itemsep=1pt,topsep=2pt]
    \item \textbf{Truthful / Correct ($y^* = 1$)}: Assigned when the candidate response conveys the factually accurate proposition supported by verified scientific consensus or gold reference answers, including valid paraphrases, acceptable scientific synonymy (e.g., specifying \textit{chewing} for mechanical breakdown, or \textit{two electrons} for one electron pair), and comprehensive explanations that accurately refute misconceptions.
    \item \textbf{Hallucinated / False ($y^* = 0$)}: Assigned when the candidate response asserts factually false or scientifically inaccurate statements, validates a misleading adversarial premise, hallucinates non-existent entities, asserts physical impossibilities (e.g., categorizing solid coal as a gas), or tautologically echoes prompt vocabulary without answering the underlying question.
\end{itemize}

\paragraph{Inter-Annotator Agreement and Adjudication.}
Annotation was conducted independently with randomized presentation. Inter-annotator reliability was assessed using Fleiss' $\kappa$ across the three annotators, yielding substantial agreement across language subsets: English ($\kappa = 0.77$), Hindi ($\kappa = 0.74$), Bengali ($\kappa = 0.75$), Tamil ($\kappa = 0.72$), and Telugu ($\kappa = 0.73$). Provisional labels were assigned by majority consensus ($2/3$ agreement). For cases exhibiting subtle scientific edge conditions or split disagreement ($<2.5\%$ of all evaluated instances), final adjudication was performed by the lead author against domain-specific scientific references to establish the final gold truth ($y^*$).

\section{Detailed Implementation Details and Hyperparameters}
\label{app:ablations}

We conduct targeted ablation studies to confirm that our empirical findings are robust to evaluation choices, probe architectures, and model hyperparameters.

\subsection{Probe Architecture and Training Setup}
\begin{itemize}[leftmargin=*,itemsep=1pt,topsep=2pt]
    \item \textbf{Probing Classifier}: All supervised hallucination detectors (SAPLMA and DRIFT) are implemented as $L_2$-regularized linear probes (Logistic Regression) optimized using L-BFGS with a convergence tolerance of $10^{-4}$ and maximum 1,000 iterations. The regularization strength is set to $C = 1.0$ by default, with stability sweeps across $C \in \{0.01, 0.1, 1.0, 10, 100\}$. Probes are evaluated across 3 fixed random seeds ($42, 123, 456$), and we report test AUROC on held-out splits.
    \item \textbf{Layer Representation Taps}: 
        In 28-layer transformer backbones, linear truthfulness decodability peaks in middle-to-late representations rather than the final unembedding layer, which is dominated by surface vocabulary selection \citep{azaria2023internal}. Layer 17 sits at $\approx 61\%$ network depth ($17/28$), where representational separation is empirically strongest:
    \begin{itemize}[nolistsep]
        \item \emph{Llama-3.2-3B} (28 layers, hidden size $d = 3072$): SAPLMA extracts activations at layer 17 ($\approx 61\%$ depth) from the final token position. DRIFT taps intermediate layers $\{8, 12, 16, 17\}$, extracting both hidden states $h^{(l)}$ and layer-transition difference vectors $\Delta h^{(l)} = h^{(l)} - h^{(l-1)}$ to represent representational trajectory dynamics.
        \item \emph{Sarvam-1} (28 layers, hidden size $d = 2048$): SAPLMA extracts final-token activations at layer 17 ($\approx 61\%$ depth). DRIFT taps layers $\{8, 12, 16, 17\}$ symmetrically.
    \end{itemize}
    \item \textbf{Unsupervised Detector Baselines}:
    \begin{itemize}[nolistsep]
        \item \emph{HaloScope-subspace}: Extracts hidden states across prompt and generation tokens, performing singular value decomposition (SVD) with subspace rank $k = 32$ to evaluate class-conditional subspace distance scoring \citep{du2024haloscope}.
        \item \emph{ICR-magnitude}: Evaluates layer-wise representation dynamics via the mean relative rate of change $\|h_{l+1}-h_l\|_2 / \|h_l\|_2$ across intermediate transformer layers (layers 10--20), implementing the representation-dynamics component without the learned MHSA-vs-FFN attribution head \citep{zhang2025icr}.
        \item \emph{Uncertainty Baselines}: Predictive entropy, length-normalized entropy, and answer log-probability are computed directly from generator output logits without additional parameters.
    \end{itemize}
    \item \textbf{Compute Infrastructure}: All model inference, label generation across 15 judges, and probe training were executed on a dedicated server equipped with $2\times$ NVIDIA RTX 6000 Ada Generation GPUs (48GB VRAM each, 96GB aggregate VRAM), an AMD EPYC 9124 16-Core Processor, and 256GB system RAM, running PyTorch 2.4 and CUDA 12.4.
\end{itemize}

\begin{figure*}[h]
    \centering
    \includegraphics[width=\linewidth]{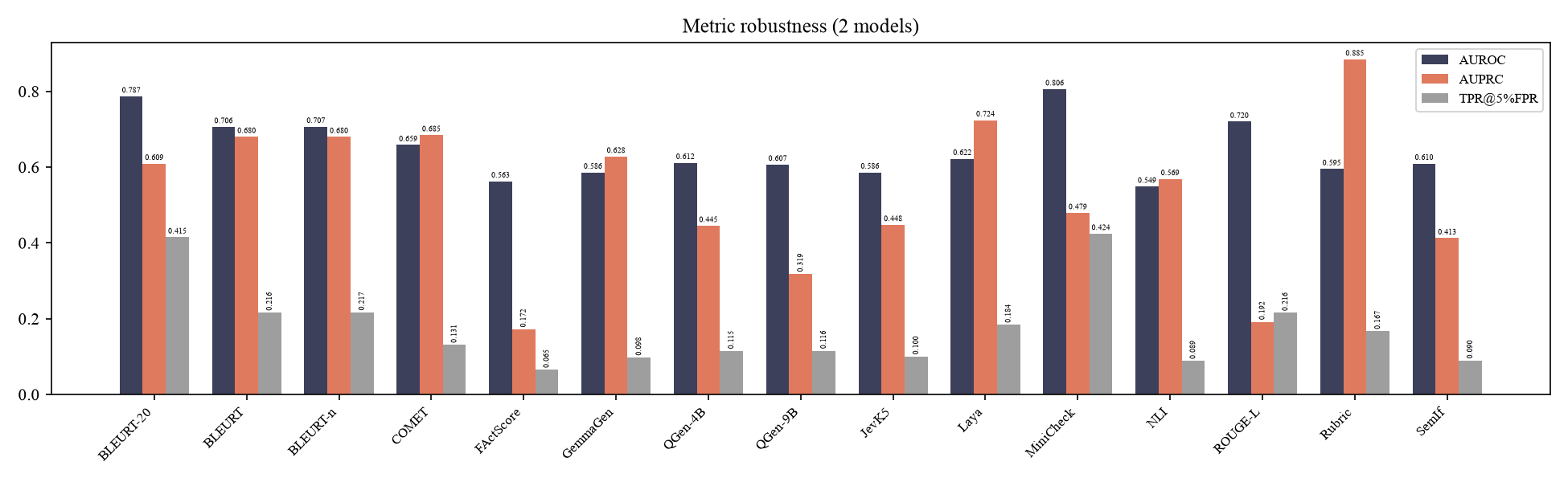}
    \caption{\textbf{The reported detector ranking depends on the evaluation metric.} Grouped bar chart comparing AUROC, AUPRC, and TPR at 5\% FPR across fifteen labelling methods. High-positivity methods are flattered by AUPRC and penalized in the low-false-alarm regime.}
    \label{fig:metric_robustness}
\end{figure*}

\subsection{Evaluation Metric and Protocol Ablations}
\paragraph{Evaluation Metric Robustness.}
Figure~\ref{fig:metric_robustness} evaluates detectors under three distinct metrics: AUROC (threshold-free ranking), AUPRC (area under precision-recall), and TPR at a fixed 5\% FPR (low-false-alarm regime). The reported ranking shifts dramatically across metrics: judges with high positivity are inflated under AUPRC but collapse under the low false-alarm requirement.

\begin{figure*}[h]
    \centering
    \includegraphics[width=\linewidth]{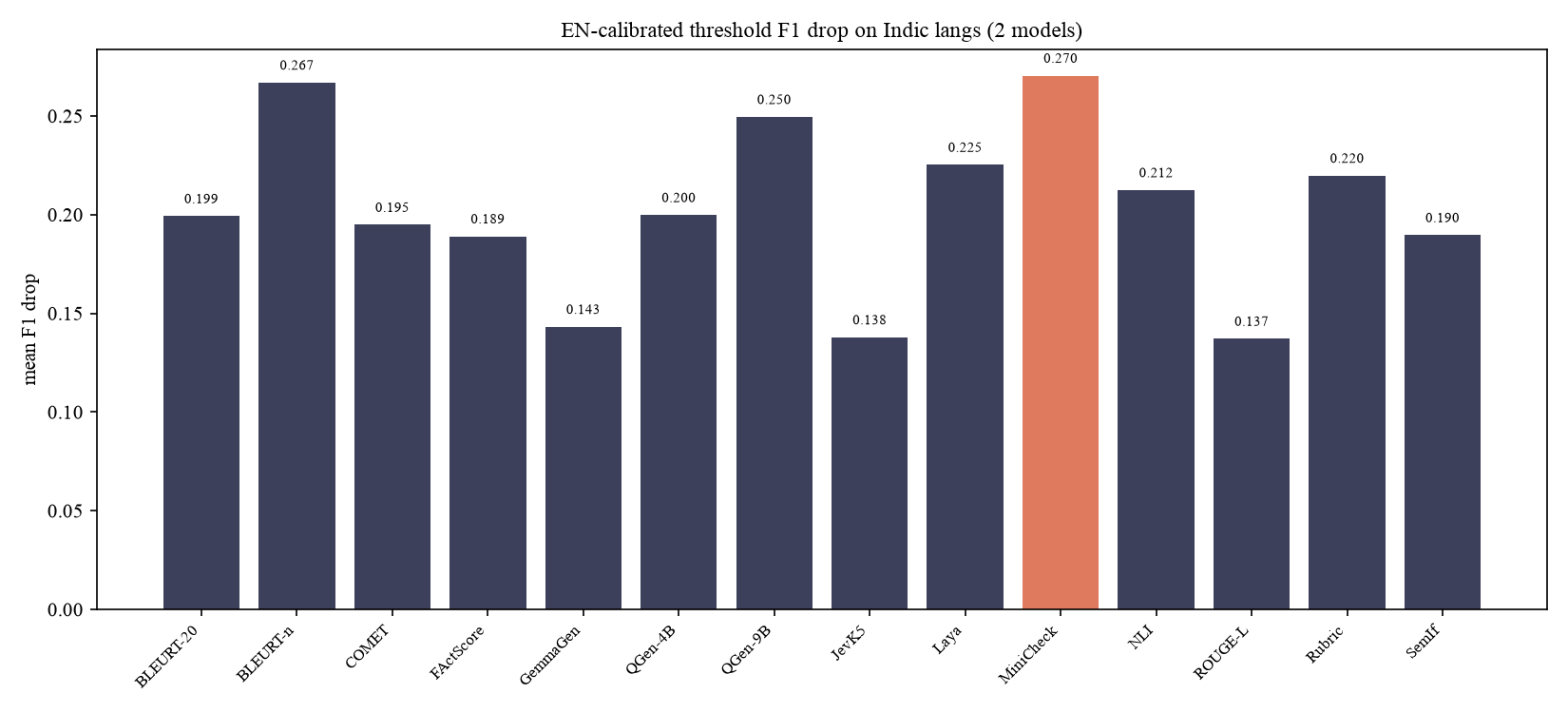}
    \caption{\textbf{Degradation of English-tuned thresholds on Indic languages.} Mean reduction in macro F1 score when operating thresholds calibrated on English are applied to Hindi, Bengali, Tamil, and Telugu without recalibration.}
    \label{fig:threshold_transfer}
\end{figure*}

\paragraph{Cross-Script Threshold Degradation.}
Figure~\ref{fig:threshold_transfer} tests the common practice of freezing operating thresholds calibrated on English and evaluating on Indic text. Every judge suffers severe degradation, with an average macro F1 drop of $0.20$ across languages (ranging from $0.14$ for ROUGE-L to $0.27$ for MiniCheck), demonstrating that operating thresholds cannot be transferred across language scripts without recalibration.

\paragraph{Additional Hyperparameter Ablations.}
\begin{itemize}[leftmargin=*,itemsep=1pt,topsep=2pt]
    \item \emph{Probe capacity}: Varying classifier regularization $C \in \{0.01, 1, 100\}$ preserves relative detector rankings throughout (Spearman rank correlation $\rho > 0.98$ across all 15 judges).
    \item \emph{Binarization rule}: Replacing the static $0.5$ threshold with a uniform median split leaves BLEURT AUROC identical at $0.706$, while ROUGE-L shifts from $0.746$ to $0.659$ across $10$ dataset-languages.
    \item \emph{Readout depth}: Intermediate-layer readout (layer 17) gains over the final layer are small in median ($+0.03$ AUROC, max $+0.38$ for ICR), showing that supervision choice decisively dominates probe layer depth.
\end{itemize}

\begin{table*}[h]
\centering
\scriptsize
\setlength{\tabcolsep}{2.0pt}
\renewcommand{\arraystretch}{0.88}
\caption{\textbf{SciQA test examples illustrating systematic supervision failures across all fifteen judges} ($0 = \text{Hallucinated / False}$, $1 = \text{Truthful / Correct}$). Predictions are from evaluated generator models (Llama-3.2-3B and Sarvam-1) on the held-out test set. Every row strictly satisfies BLEURT $\ne$ Gold and 15-judge majority vote $\ne$ Gold against human gold truth.}
\label{tab:sciq_case_studies_table}
\adjustbox{max width=\textwidth}{%
\begin{tabular}{@{}lp{5.6cm}p{4.3cm}p{2.5cm}|ccc|cc|ccc|cccc|ccc|c@{}}
\toprule
& & & & \multicolumn{3}{c|}{\textbf{Learned}} & \multicolumn{2}{c|}{\textbf{Surface}} & \multicolumn{3}{c|}{\textbf{Entailment}} & \multicolumn{4}{c|}{\textbf{Generative}} & \multicolumn{3}{c|}{\textbf{Decision}} & \textbf{Gold} \\
\textbf{Category} & \textbf{Question} & \textbf{Model Prediction} & \textbf{Correct / Reference} & \rotatebox{90}{BL} & \rotatebox{90}{BL-n} & \rotatebox{90}{B-20} & \rotatebox{90}{COM} & \rotatebox{90}{RGE} & \rotatebox{90}{NLI} & \rotatebox{90}{MNC} & \rotatebox{90}{FCS} & \rotatebox{90}{Q-4B} & \rotatebox{90}{Q-9B} & \rotatebox{90}{GEM} & \rotatebox{90}{RUB} & \rotatebox{90}{JEV} & \rotatebox{90}{SIF} & \rotatebox{90}{LAY} & \rotatebox{90}{\textbf{Truth}} \\
\midrule
\textbf{Lexical Overlap} & \mbox{}\telugutxt{పెద్ద ప్రేగులలో నివసించే ట్రిలియన్ల బ్యాక్టీరియాలను ఏమంటారు?} \newline \textit{(What are the trillions of bacteria in the large intestine called?)} & \mbox{}\telugutxt{అవి పెద్ద ప్రేగులో నివసించే ట్రిలియన్ల బ్యాక్టీరియాను సూచిస్తాయి.} \newline \textit{(They refer to the trillions of bacteria living in the large intestine.)} & \mbox{}\telugutxt{బాక్టీరియా వృక్షజాలం} \newline \textit{(bacteria flora)} & 1 & 1 & 0 & 1 & 0 & 1 & 0 & 1 & 1 & 0 & 1 & 1 & 1 & 1 & 1 & \textbf{0} \\
\textbf{Misconception} & What is the physical breakdown of chunks of food into smaller pieces? & The physical breakdown of chunks of food into smaller pieces is called chewing. & mechanical digestion & 0 & 0 & 0 & 1 & 0 & 1 & 0 & 1 & 0 & 0 & 1 & 1 & 0 & 0 & 1 & \textbf{1} \\
\textbf{Partial Correctness} & \mbox{}\hinditxt{उन श्वेत रक्त कोशिकाओं का नाम बताइए जो केशिका दीवारों के माध्यम से निचोड़ सकती हैं \dots} \newline \textit{(Name the white blood cells that squeeze through capillary walls \dots)} & \mbox{}\hinditxt{उत्तरः फागोसाइट्स, विशेष रूप से न्यूट्रोफिल और मैक्रोफेज \dots} \newline \textit{(Answer: phagocytes, especially neutrophils and macrophages \dots)} & \mbox{}\hinditxt{न्यूट्रोफिल} \newline \textit{(neutrophils)} & 0 & 0 & 0 & 0 & 0 & 1 & 0 & 0 & 1 & 0 & 1 & 1 & 1 & 1 & 1 & \textbf{1} \\
\textbf{Real-World Premise} & \mbox{}\tamiltxt{திட நிலக்கரி, திரவ பெட்ரோலியம் மற்றும் திரவ இயற்கை எரிவாயு அனைத்தும் என்ன வகைகள்?} \newline \textit{(Solid coal, liquid petroleum and liquid natural gas are all what types?)} & \mbox{}\tamiltxt{திட நிலக்கரி, திரவ பெட்ரோலியம் மற்றும் திரவ இயற்கை எரிவாயு மூன்று வகையான எரிவாயுகள் ஆகும்.} \newline \textit{(... are three types of gases.)} & \mbox{}\tamiltxt{புதைபடிவ எரிபொருள்கள்} \newline \textit{(fossil fuels)} & 1 & 1 & 0 & 1 & 0 & 0 & 0 & 0 & 1 & 0 & 1 & 1 & 1 & 1 & 1 & \textbf{0} \\
\textbf{Numerical Reasoning} & Each hydrogen atom follows the octet rule with how many pairs of electrons? & The hydrogen atom follows the octet rule with two electrons. & one & 0 & 0 & 0 & 0 & 0 & 0 & 0 & 0 & 0 & 0 & 0 & 1 & 0 & 0 & 0 & \textbf{1} \\
\textbf{Logical Reasoning} & \mbox{}\tamiltxt{தொடுதல், இருமல் அல்லது தும்மல் ஆகியவை டெட்டனஸ் போன்ற நோய்களை ஏற்படுத்தும் எந்த நுண்ணுயிரிகளை நேரடியாகப் பரப்புகின்றன?} \newline \textit{(Touching/coughing/sneezing directly spread what tetanus-causing microbes?)} & \mbox{}\tamiltxt{\dots நேரடித் தொடுதல், இருமல் அல்லது தும்மல் மூலம் பரவுகின்றன.} \newline \textit{(\dots spread via direct touch, cough or sneeze.)} & \mbox{}\tamiltxt{பாக்டீரியா} \newline \textit{(bacteria)} & 1 & 1 & 0 & 1 & 0 & 1 & 0 & 0 & 0 & 0 & 1 & 1 & 1 & 1 & 1 & \textbf{0} \\
\textbf{Multilingual Failure} & \mbox{}\bengalitxt{ক্যাফেইন এবং অ্যালকোহল কোন ধরনের মাদকের দুটি উদাহরণ?} \newline \textit{(Caffeine and alcohol are two examples of what type of drug?)} & \mbox{}\bengalitxt{উত্তর: ক্যাফিন এবং অ্যালকোহল} \newline \textit{(Answer: caffeine and alcohol.)} & \mbox{}\bengalitxt{সাইকোঅ্যাকটিভ} \newline \textit{(psychoactive)} & 1 & 1 & 0 & 1 & 0 & 1 & 0 & 0 & 0 & 1 & 1 & 1 & 1 & 1 & 1 & \textbf{0} \\
\bottomrule
\end{tabular}}
\end{table*}

\section{Qualitative Failure Case Studies on SciQ}
\label{app:sciq_case_studies}

To complement the TruthfulQA qualitative audit in Section~\ref{sec:qual_english_failures} (Table~\ref{tab:tqa_case_studies}), Table~\ref{tab:sciq_case_studies_table} examines seven representative failure modes on the science question answering benchmark SciQ across English and four Indic languages. Every presented case satisfies the strict audit criterion where both BLEURT and the 15-judge majority vote diverge from human-verified gold ground truth ($y^*$). All seven examples are drawn exclusively from the held-out SciQ test evaluation sheets, covering all five benchmark languages (English, Hindi, Bengali, Tamil, and Telugu).

\section{Detailed Cross-Ground-Truth Transfer}
\label{app:xgt_full}



\begin{figure}[h]
    \centering
    \includegraphics[width=\linewidth]{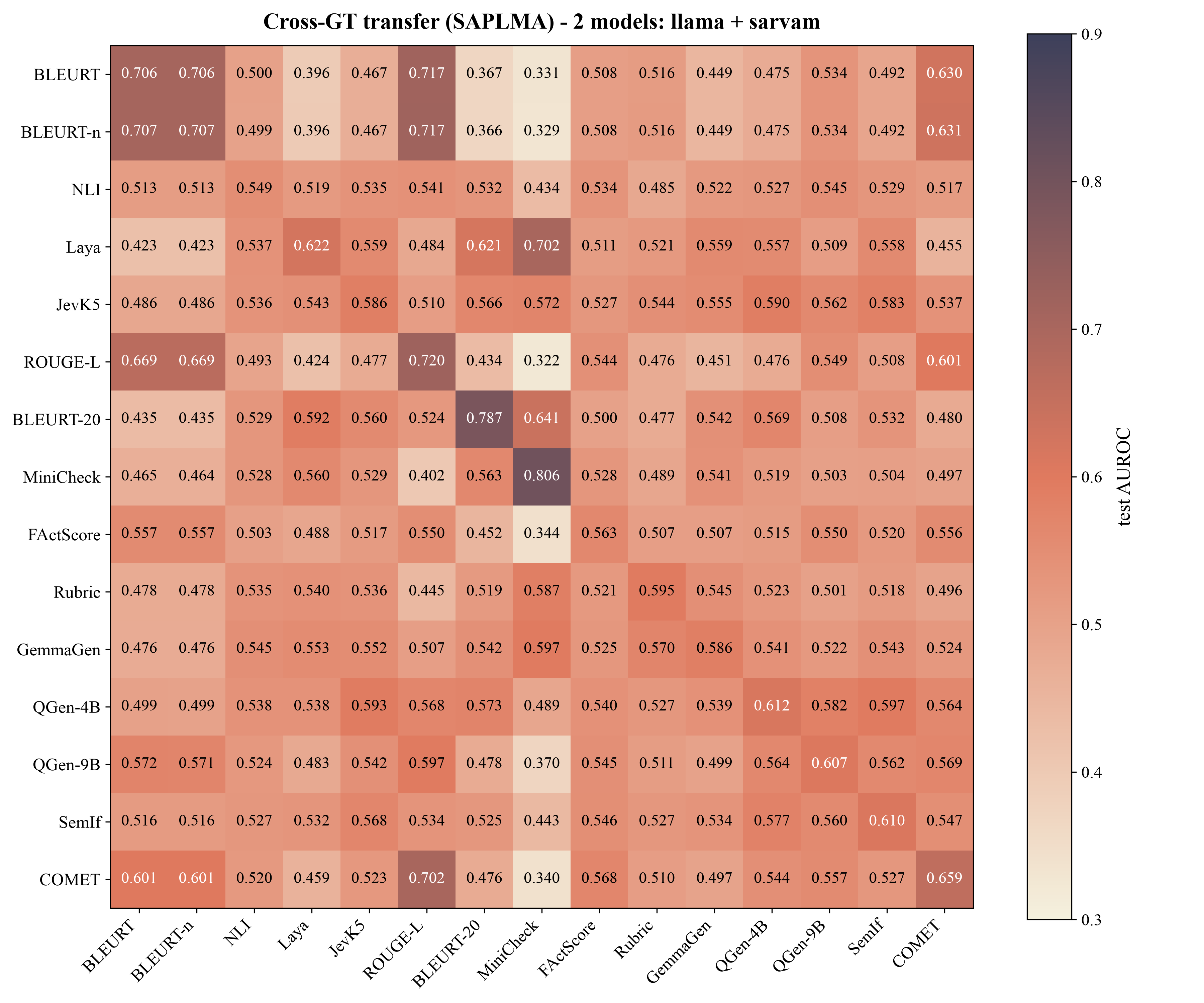}
    \caption{\textbf{Cross-ground-truth transferability under SAPLMA probes.} Evaluated across $20$ experimental conditions. The in-judge diagonal reaches an average AUROC of $0.641$, whereas off-diagonal cross-judge transfer collapses to $0.524$ (chance level $\approx 0.500$).}
    \label{fig:cross_gt_saplma}
\end{figure}

Figure~\ref{fig:cross_gt_saplma} reports the transfer matrix for single-layer SAPLMA probes based on final-token representations. Paralleling the multi-layer DRIFT probes in the main paper (Figure~\ref{fig:cross_gt_matrix}), SAPLMA achieves a diagonal average AUROC of $0.641$, while off-diagonal transfer degrades to $0.524$ (chance level $\approx 0.500$). Non-trivial transfer ($>0.60$ AUROC) concentrates overwhelmingly within the surface-similarity cluster (BLEURT, BLEURT-n, COMET, ROUGE-L), alongside isolated MiniCheck pairings, suggesting that shared lexical overlap and score alignment, rather than universal factual ground truth, may account for localized transfer clusters.

\section{Fine-Grained Preliminary Experiments}
\label{app:detailed_baselines}

This section decomposes benchmark detector performance across individual languages, generator architectures, and benchmark datasets, expanding the aggregate results reported in Table~\ref{tab:methods}. Tables~\ref{tab:breakdown_lang}, \ref{tab:breakdown_model}, and \ref{tab:breakdown_dataset} present the three primary breakdowns from the Preliminary Experiments. Tables~\ref{tab:judges_lang} and \ref{tab:judges_models} provide corresponding decompositions for each supervision judge across languages and generator backbones.

\begin{figure*}[h]
    \centering
    \includegraphics[width=\linewidth]{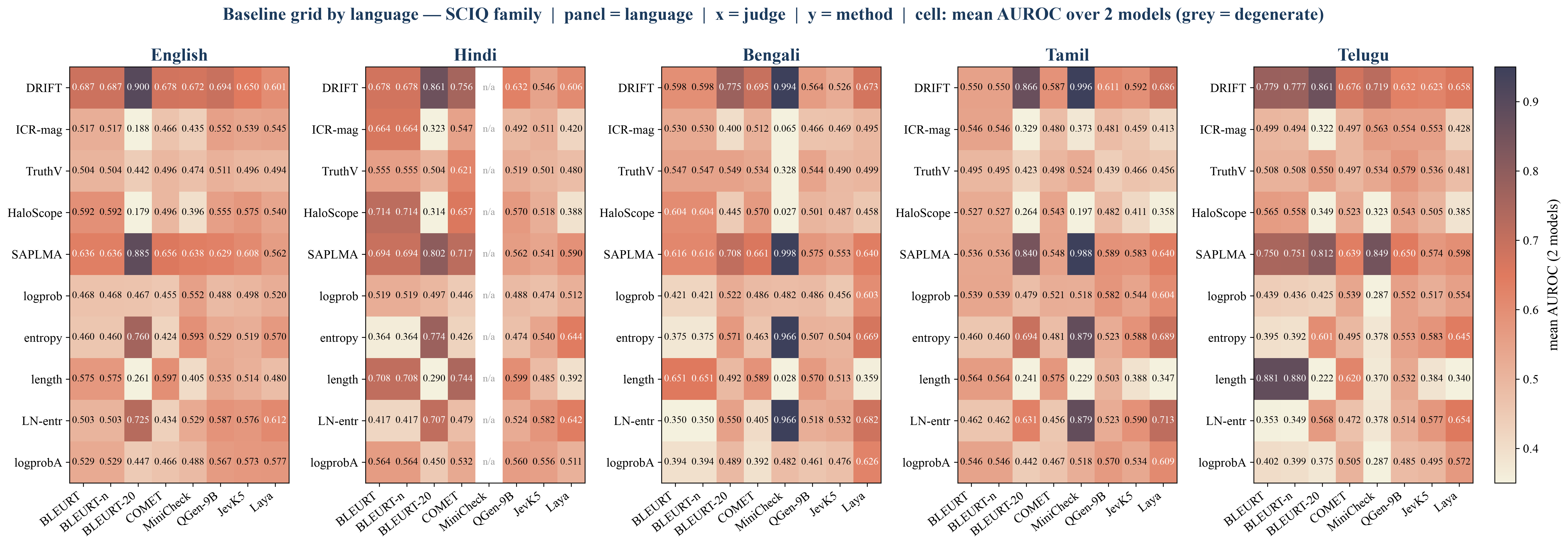}
    \caption{\textbf{Language-wise baseline performance on SciQ.} Fine-grained evaluation grid across English, Hindi, Bengali, Tamil, and Telugu, depicting mean AUROC over two generator models across eight representative supervision judges and ten detection methods.}
    \label{fig:gridlang_sciq}
\end{figure*}

Figure~\ref{fig:gridlang_sciq} displays the fine-grained language-wise baseline grid for the SciQ dataset family across 2 generator models, 8 representative supervision judges spanning all five technical mechanisms (BLEURT, BLEURT-n, BLEURT-20 for learned similarity; COMET for translation quality; MiniCheck for textual entailment; QGen-9B for generative verdicts; and JevK5, Laya for decision agents), and 10 detection methods comprising supervised probes (DRIFT, SAPLMA), unsupervised representation-dynamics baselines (ICR-mag, HaloScope-subspace, TruthV), and output uncertainty/heuristic estimators (logprob, entropy, LN-entropy, logprob-ans, length). Paralleling TruthfulQA, supervised probes lead discrimination while unsupervised uncertainty baselines hover near chance across all five languages.

\begin{table}[h]
\centering
\scriptsize
\setlength{\tabcolsep}{2.5pt}
\caption{\textbf{Detector performance by language on Preliminary Experiment} Test AUROC averaged across all 15 judges and 2 models. Best per column in bold.}
\label{tab:breakdown_lang}
\begin{tabular}{@{}lrrrrrr@{}}
\toprule
Method & Eng & Hin & Ben & Tam & Tel & Avg \\
\midrule
DRIFT & \textbf{0.681} & \textbf{0.671} & \textbf{0.656} & \textbf{0.639} & \textbf{0.672} & \textbf{0.664} \\
SAPLMA & 0.653 & 0.635 & 0.645 & 0.615 & 0.645 & 0.639 \\
ICR-mag. & 0.493 & 0.502 & 0.486 & 0.484 & 0.477 & 0.488 \\
HaloScope & 0.514 & 0.538 & 0.479 & 0.469 & 0.482 & 0.496 \\
Norm. Entropy & 0.567 & 0.519 & 0.553 & 0.576 & 0.553 & 0.554 \\
Answer Log-P & 0.543 & 0.514 & 0.531 & 0.525 & 0.499 & 0.522 \\
Length Ctrl & 0.519 & 0.551 & 0.482 & 0.471 & 0.498 & 0.504 \\
\bottomrule
\end{tabular}
\end{table}

\begin{table}[h]
\centering
\scriptsize
\setlength{\tabcolsep}{2.2pt}
\caption{\textbf{Detector performance by generator architecture on on Preliminary Experiment} Test AUROC averaged across 15 judges and 10 dataset-languages. Best per column in bold.}
\label{tab:breakdown_model}
\begin{tabular}{@{}lrrrrrr@{}}
\toprule
Method & Llama & Sarvam & Avg \\
\midrule
DRIFT & \textbf{0.609} & \textbf{0.715} & \textbf{0.664} \\
SAPLMA & 0.598 & 0.677 & 0.639 \\
ICR-mag. & 0.503 & 0.475 & 0.488 \\
HaloScope & 0.521 & 0.473 & 0.496 \\
Norm. Entropy & 0.565 & 0.544 & 0.554 \\
Answer Log-P & 0.543 & 0.504 & 0.522 \\
Length Ctrl & 0.522 & 0.488 & 0.504 \\
\bottomrule
\end{tabular}
\end{table}

\begin{table}[h]
\centering
\scriptsize
\setlength{\tabcolsep}{5pt}
\caption{\textbf{Detector performance by benchmark dataset on Preliminary Experiment} Test AUROC averaged across 15 judges, 2 models, and 5 languages. Best per column in bold.}
\label{tab:breakdown_dataset}
\begin{tabular}{@{}lrrr@{}}
\toprule
Method & TruthfulQA & SciQ & Avg \\
\midrule
DRIFT & \textbf{0.684} & \textbf{0.642} & \textbf{0.664} \\
SAPLMA & 0.656 & 0.621 & 0.639 \\
ICR-mag. & 0.499 & 0.477 & 0.488 \\
HaloScope & 0.506 & 0.486 & 0.496 \\
Norm. Entropy & 0.567 & 0.540 & 0.554 \\
Answer Log-P & 0.534 & 0.510 & 0.522 \\
Length Ctrl & 0.502 & 0.507 & 0.504 \\
\bottomrule
\end{tabular}
\end{table}

\begin{table}[h]
\centering
\scriptsize
\setlength{\tabcolsep}{4pt}
\caption{\textbf{Supervision judge performance broken down by generator architecture.} Test AUROC averaged across the preliminary experiment. Best per column in bold. The Avg column reports the macro-average across the two generator backbones.}
\label{tab:judges_models}
\begin{tabular}{@{}lrrr@{}}
\toprule
Judge & Llama & Sarvam & Avg \\
\midrule
BLEURT     & 0.560 & \textbf{0.671} & \textbf{0.616} \\
BLEURT-n   & 0.560 & \textbf{0.671} & 0.615 \\
BLEURT-20  & 0.577 & 0.511 & 0.544 \\
COMET      & 0.555 & 0.599 & 0.577 \\
ROUGE-L    & \textbf{0.615} & 0.561 & 0.588 \\
NLI        & 0.517 & 0.523 & 0.520 \\
MiniCheck  & 0.520 & 0.523 & 0.522 \\
FActScore  & 0.533 & 0.547 & 0.540 \\
QGen-4B    & 0.556 & 0.524 & 0.540 \\
QGen-9B    & 0.557 & 0.557 & 0.557 \\
GemmaGen   & 0.539 & 0.517 & 0.528 \\
JevK5      & 0.546 & 0.515 & 0.531 \\
SemIf      & 0.583 & 0.531 & 0.557 \\
Rubric     & 0.522 & 0.533 & 0.527 \\
Laya       & 0.538 & 0.520 & 0.529 \\
\bottomrule
\end{tabular}
\end{table}

\begin{table}[h]
\centering
\scriptsize
\setlength{\tabcolsep}{2.5pt}
\caption{\textbf{Supervision judge performance broken down by language.} Test AUROC averaged across all Preliminary benchmarks. Chance level is $0.500$. Best per column in bold. The Avg column reports the macro-average across the five languages.}
\label{tab:judges_lang}
\begin{tabular}{@{}lrrrrrr@{}}
\toprule
Judge & Eng & Hin & Ben & Tam & Tel & Avg \\
\midrule
BLEURT     & \textbf{0.621} & \textbf{0.654} & 0.586 & 0.594 & \textbf{0.623} & \textbf{0.616} \\
BLEURT-n   & 0.620 & \textbf{0.654} & 0.586 & 0.593 & 0.622 & 0.615 \\
BLEURT-20  & 0.551 & 0.558 & 0.562 & 0.526 & 0.525 & 0.544 \\
COMET      & 0.587 & 0.625 & 0.562 & 0.548 & 0.562 & 0.577 \\
ROUGE-L    & 0.569 & 0.632 & \textbf{0.590} & \textbf{0.604} & 0.515 & 0.582 \\
NLI        & 0.550 & 0.508 & 0.524 & 0.510 & 0.507 & 0.520 \\
MiniCheck  & 0.516 & 0.538 & 0.509 & 0.541 & 0.522 & 0.525 \\
FActScore  & 0.566 & 0.550 & 0.543 & 0.515 & 0.525 & 0.540 \\
QGen-4B    & 0.568 & 0.535 & 0.535 & 0.524 & 0.539 & 0.540 \\
QGen-9B    & 0.579 & 0.556 & 0.534 & 0.537 & 0.579 & 0.557 \\
GemmaGen   & 0.544 & 0.512 & 0.522 & 0.536 & 0.525 & 0.528 \\
JevK5      & 0.570 & 0.524 & 0.517 & 0.507 & 0.536 & 0.531 \\
SemIf      & 0.587 & 0.565 & 0.566 & 0.531 & 0.538 & 0.557 \\
Rubric     & 0.547 & 0.504 & 0.525 & 0.532 & 0.528 & 0.527 \\
Laya       & 0.536 & 0.507 & 0.542 & 0.532 & 0.529 & 0.529 \\
\bottomrule
\end{tabular}
\end{table}

\FloatBarrier

\end{document}